\documentclass[11pt]{article}

\usepackage{graphicx}
\usepackage{amsfonts}
\usepackage{float}
\usepackage{amsmath, amssymb, latexsym, amscd, amsthm, epsfig}
\usepackage{makeidx}
\title{\textbf{The principle of cognitive action} \\  {\Large Preliminary experimental analysis} }
\author{     \vspace{1cm}   \\  { Marco Gori, Marco Maggini, Alessandro Rossi}   \\  \vspace{0.5cm} \\ {  Department of Information Engineering and Mathematical Sciences, } \\ \vspace{0.1cm} \\ {University of Siena, Italy  }       \\ \vspace{1cm}     }
\date{   May 20, 2015  \\ \vspace{0.1cm}}

\def\A{\mathbf A}
\def\B{\mathbf B}

\def\W{\mathbf W}
\def\a{\mathbf a}

\def\g{\mathbf g}

\def\u{\mathbf u}
\def\x{\mathbf x}
\def\y{\mathbf y}

\def\z{\mathbf z}

\def\cD{\mathcal D}

\def\cL{\mathcal L}

\def\cP{\mathcal P}

\def\cS{\mathcal S}
\def\cT{\mathcal T}

\def\cX{\mathcal X}

\def\R{\mathbb{R}}

\def\x{\mathbf{x}}
\def\y{\mathbf{y}}
\def\by{\bar{f}}
\def\u{\mathbf{u}}
\def\W{\mathbf{W}}

\def\inn{\! \in \!}
\def\ft{\! : \!}
\def\ug{\! = \!}

\def\g{\gamma}

\def\t{\theta}
\def\l{\lambda}
\def\z{\zeta}
\def\b{\beta}
\def\q{\tau}
\def\a{\alpha}

\begin{document}

\maketitle

%\printindex

\begin{abstract}

In this document we shows a first implementation and some preliminary results of a new theory, facing Machine Learning problems in the frameworks of Classical Mechanics and Variational Calculus. We give a general formulation of the problem and then we studies basic behaviors of the model on simple practical implementations.

\end{abstract}

\newpage 

\tableofcontents

\newpage

% % % % % % INTRO
\section{Introduction}

Many real world phenomena could be interpreted in an on-line scenario within Machine Learning theory. In long-life learning problems, various approaches have been developed to deal with the large amounts of data and the exploitation of their time correlation. Usually, some difficulties arise in the storage of data and in going after the intrinsic information coming from data during time. These aspects could suggest a more natural approach to learning, in a joint theory among Mechanics, Variational Calculus and Statistics. However, we postpone a deeply analysis of these ideas at a theoretical level, focusing on a first practical implementation, to create a connection between this new ideas and some applications on existing structures, which could be useful in these preliminary steps. 

We briefly introduce basics ideas of this theory, first formulated in~\cite{FrandinaGLMM13}. The concept of dissipation is well formulated in~\cite{Betti2015},whereas an summing up of this report and an experimental analysis on standard benchmark can be find in~\cite{Gori201672}. A further theoretical abstraction applied to similar environment is proposed in~\cite{Maggini2016}, In this document, we will give a slightly theoretical formulation in order to allow us to go straight to the practical implementation issues. We study the meaning of the model parameters on artificial problems, using a linear function, and  then we try some experiments  with simple Neural Networks.

An important result is the fact that we can choose arbitrarily the memory of our system by a parameter, avoiding the hardware memory storage problems. Indeed, the trend of the system is influenced by the information coming from each example during an adjustable interval of time.

% % % % % % FORMULATION
\section{Formulation of the problem}

We study the case in which we want to learn a function $f$ that aims to represent the behavior of a features representation $\u$ of the spatio-temporal domains $\cD$. If we assume the temporal domain 
$\cT \! \! = \left[ 0, \infty \right) $
 and
 $\cX \! \! \subseteq {\R}^D $
 then we have
 $\cD = \cT \! \times \! \cX$ 
 and
 $ \u  \ft  \cD \rightarrow {\R}^d $ 
so that $f$ has input $\u$ and depends on a set of weights $\W$, like for example if $f$ is described by an Artificial Neural Network. The problem consits on learning the parameters $\W$ under some assumption, i.e. by minimizing a cost functional $\cL$ composed by a penalty term and a regularization one.  Since the weights have to be learned as time goes by, we have that  $\W$  depends on time and we write $f \ug f(\u(t,\x),\W(t))$. In the classical approach the penalty term impose a coherence w.r.t. the Training Set, whereas  the regularization term impose the norm of the parameters to be small, so as to $f$ be a smooth function. If we want the process of learning itself to be smooth, we could requires some kind of regularization in the changing of $\W$ during time. We shall see in next sections how this idea could be studied in a \emph{Physics-like} approach.

% % % % % % COST FUNC
\section{Construction of Cost Functional}

In a classical learning problem we have to minimize a cost functional $\cL$ w.r.t. $\W(t)$. The functional is composed by a penalty term and a regularization one. The penalty term is calculated on every supervised example, over a training set $\cP \ug {\{(u_k , {\by}_k)\}}_{k=1}^l$, by a loss function $V \ug f(\u(t,\x),\W(t))$ which is the summation over $\cP$ :
$$
V  \left( \u(t,\x),\W(t) \right) = \sum_{k=1}^{l}  \overline{V} \left( f(\u(t,\x),\W(t))\:  , \:{\by}_k \right)
$$
where $ \overline{V}$ could be for example the quadratic function $\overline{V} ( f , {\by}_k ) \ug \frac{1}{2} {(f-{\by}_k)}^2$ ( where $f$ means $f(\u(t,\x),\W(t))$ from now on ).
Since the examples are presented in time, if $t_k$ is the instant of time in which the couple $(u_k , {\by}_k)$ is provide, we can write $V$ as
$$
V  \left( \u(t,\x),\W(t) \right) = \sum_{k=1}^{l}  \overline{V} \left( f\:  , \:{\by}_k \right) \cdot H(t-t_k).
$$

In a physics-like approach we can look at the term $V$ as the \emph{potential energy}, so as the total energy of the system $L$ is composed by $K \! + \!V$ where $K$ is the \emph{kinetic energy} . Then in our formulation we write
\begin{equation}\label{L}
L \ug K\! + \!\g V
\end{equation}
where $\g$ represent a \emph{regularization parameter} (which includes the classical case when $\g \ug 1$ ). We can write $K$ as:
$$
K= \sum_{i=1}^m \mu_i {\dot{\omega}_i}^2
$$
where $m$ is the total number of weights in $\W$, and $\mu_i$ represent the \emph{mass} of each weight, (i.e. an additional parameters which can be use to choose the \emph{inertia} of each weight). Since $\dot{\omega}_i \ug \frac{d}{dt} \omega_i$ we can replace $D \ug \frac{d}{dt}$ with a general differential operator $T$. Now we can finally write our cost functional as
\begin{equation}\label{S}
{\cS}_{\g} = \int_{0}^{t_e} \psi(t) \cdot L \; dt.
\end{equation}
where $\psi$ is a suitable dissipation function, that we take as $\psi(t) \ug e^{\t t}$.

% % % % % %
\section{First Application}

% % % \input ./Subsections/FirstOrderOperator
\subsection{First Order Operator}% $T \ug \alpha_0 \! + \! \alpha_1 D$}
In our first application of this theoretical framework we analyze the simple case in which $f$ is a linear function of one single real variable $f \! : \!  \R \rightarrow \R $ and $ f \ug y u \! + \! b $ where $u \ug u(t,x(t))) \ug x(t) $ . In this framework we want to learn the two weights $y,b$. The problem can be formulated as

$$
y^{*} = arg \min_{y \in \R} \int_{0}^{t_e} \psi ( t ) \cdot L \; dt
$$
and analogous formula hold for $b$.

We start with the case $T \ug \alpha_0 \! + \! \alpha_1 D$. By the application of the \emph{Eulero-Lagrange} equation we have to solve the second order linear differential equation:

\begin{equation}\label{diff}
\ddot{y} + \theta \dot{y} + \beta y - \frac{\gamma}{\mu \alpha_1^2 } \sum_{k =1}^l (u_k y_k + b_k - \by_k)u_k \cdot \delta (t-t_k) =0 
\end{equation}

where $ \beta \ug \frac{\alpha_0 \alpha_1 \theta - \alpha_0^2}{\alpha_1^2}$ . The solution is then composed by a term given from the homogeneous solution $y^o(t)$ plus a term given from the impulsive response $y^F(t)$. Then we have\footnote{see section \ref{appendix} for details about practical calculation}:
\begin{equation}\label{upfo}
y(t)=y^o(t) + y^F(t) = y^o(t) + \frac{\gamma}{\mu \, \alpha_1^2 } \sum_{k =1}^l \z_k \cdot g (t-t_k)
\end{equation}
where we posed $\z_k=(u_k y_k + b_k - \by_k)u_k$. For the bias $b$ we have an analogous solution except for the element $\z_k$, which represent $\frac{\partial \overline{V}}{\partial y}$ and in the correspondent formula for $b$ is $\z_k \ug u_k y_k \!+ \! b_k \! - \! \by_k$.

For the stability of the system during time, we have to impose the \emph{Routh-Hurwitz conditions}, which lead to the relation $\t>\a_0/\a_1$.

\subsubsection{Experimental results}\label{fooe}

A first implementation (in MatLab) of our theoretical results is in the simple case in which we want to approximate a linear function in an interval $[a,b]=[-1,1]$ of the real axis. We assume that the examples on the training set are equally spaced in time by a factor $\q$ with $t_0\ug 0$. This allow us to use a discretization of (\ref{upfo}) for the computing of the evolution of the system, as you can see in Section(\ref{discretization}). We also consider an equally-spaced subdivision of our interval that we cover forward and backward, i.e. we move from $a$ to $b$ and viceversa, so as to guarantee time correlation among the examples. We assign to every point $u_k$ a target  $\by_k=2  \! \cdot \! u_k \! -\! 1$, so we desired $y(t) \! \rightarrow \! 2$ and $b(t)\! \rightarrow \! -1$ after some epochs. For start, we feed the system with a total supervised training set. We studied some results on this first implementation w.r.t. the parameters $\t, \a_0, \a_1, \g , \mu , \tau$.

\begin{figure}[H]
\hspace{-0.8cm}
\includegraphics[scale=0.7]{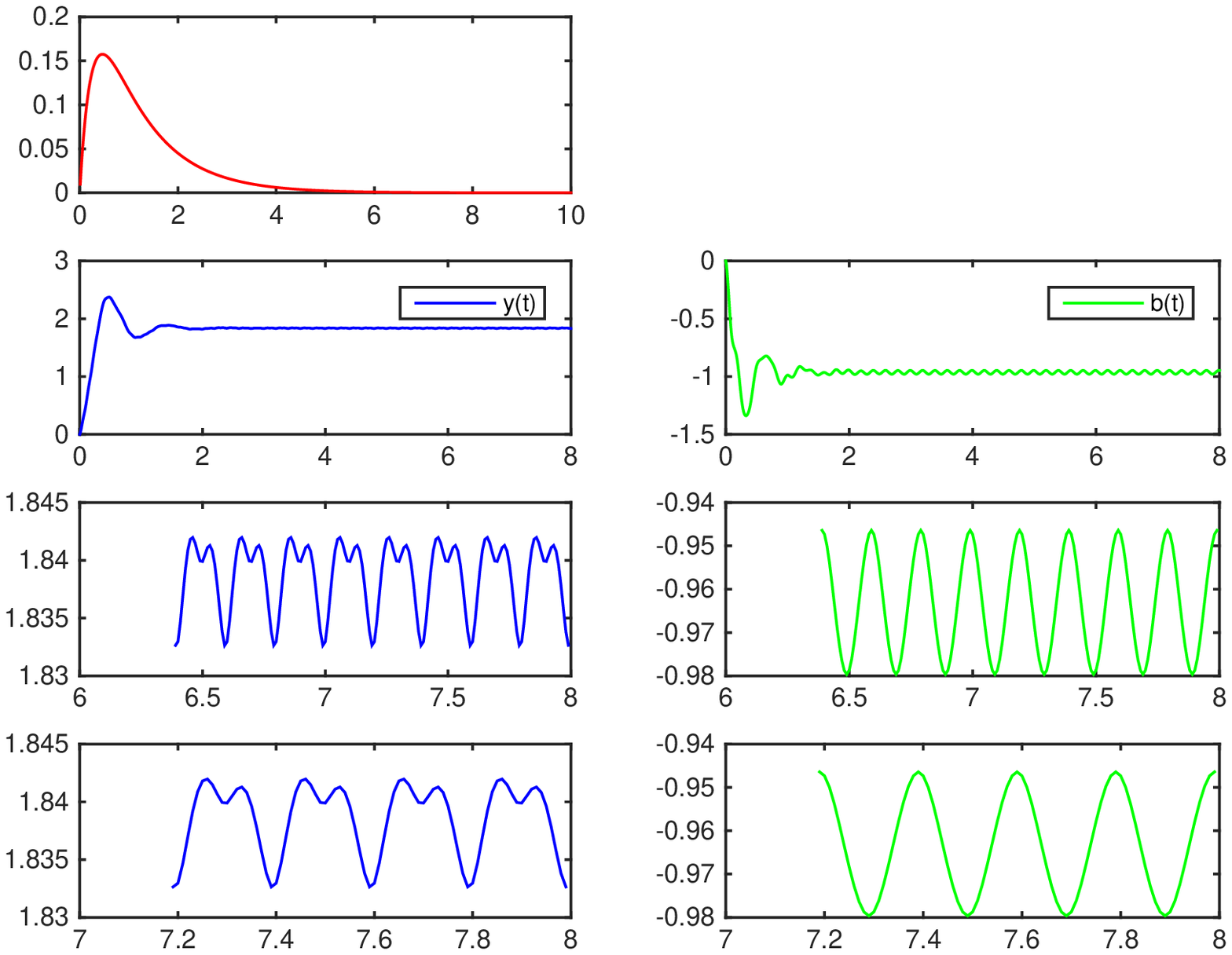}
\caption{$\g \ug -1 \, , \; \mu \ug 1 \, , \; \t \ug 5 \, , \; \a_0 \ug 1 \, , \;  \a_1 \ug 1 \, , \;   y(0)\ug y'(0)\ug b(0)\ug b'(0) \ug0 \, , \: \q \ug 0.01$, iterations $\ug40$. \label{f1}}
\end{figure}

If we start our study for a simple case we have a behavior as shown in Fig.\ref{f1} . We set $\g \ug-1 \, , \;  \mu \ug 1 \, , \; \t \ug 5 \, , \; \a_0 \ug 1 \, , \;  \a_1\ug 1  \, , \;  y(0)\ug y'(0)\ug b(0)\ug b'(0)=0$. We repeat the training set for a total of $40$ iterations.  From the top to the bottom of the figure we can find :
\begin{itemize}
\item the plot of the impulse response $g$(red line)
\item  the plot of $y$(blue line) and $b$(green line)
\item the plot of the last 20\% of update of the weights
\item the plot of the last 10\% of update of the weights
\end{itemize}
Each plot have the same scaling referred to the time $t$, so as in the graphs is reported the evolution of the system w.r.t. real time (we can think in seconds). The width of each plot depends on both the number of iterations and the parameter $\q$, but each plot has the same scale in $seconds$. As we can see in Fig.\ref{f1} the weights have an initial oscillation and then assume a cyclic behavior with a smaller amplitude oscillation when the system is a steady state, near the desired value $2,-1$ respectively. 

\newpage
In the remainder of this section we studied some variations produced by each parameters on the behavior of the weights. 

\begin{description}

\item{\bfseries{Parameter} $\mathbf{\g}$}

 In our formulation of the theory this parameters only determine the sign between the two terms $K$ and $V$, so we only explore the set $\{-1,1\}$ for this one. Because of some correlation with the \emph{gradient descent}, we expect our weights to have a divergent trend when $\g=1$. Our experimental results confirm this, as we can see in Fig.\ref{f2}, where we can notice the trend of the weights after just $5$ iterations.
\begin{figure}[H]
\hspace{-0.8cm}
\includegraphics[scale=0.7]{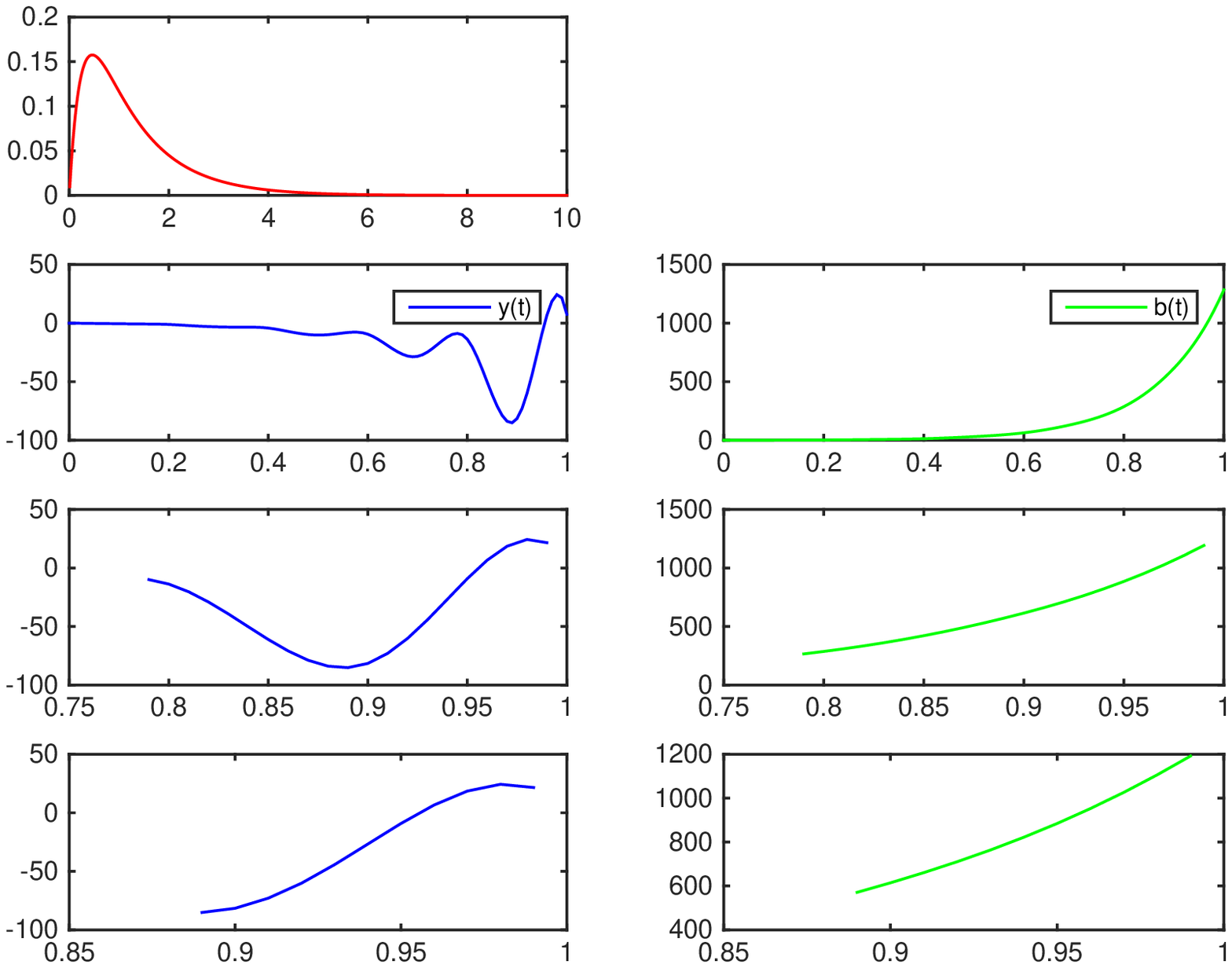}
\caption{$\g \ug 1 \, , \; \mu \ug 1 \, , \; \t \ug 5 \, , \; \a_0 \ug 1 \, , \;  \a_1 \ug 1 \, , \;   y(0)\ug y'(0)\ug b(0)\ug b'(0) \ug0 \, , \: \q \ug 0.01$, iterations $\ug 5$. \label{f2}}
\end{figure}
We tried to vary the other parameters in order to make the weights converge to the desired value also in the case $\g \ug 1$. We obtained converge by imposing a strong regularization by the differential operator ($\a_0 \, , \; \a_1 > 10$), but is difficult to find a correlation between the final values and the desired ones(Fig.\ref{f3} ). The same result ca be achieved with $\mu >30$ or $\t >120$ and also by increasing the parameter $\q$. All these adjustments on the other parameters represent the imposition of a strong regularization on the system, which drive all the weights to zero.

\begin{figure}[H]
\hspace{-0.8cm}
\includegraphics[scale=0.7]{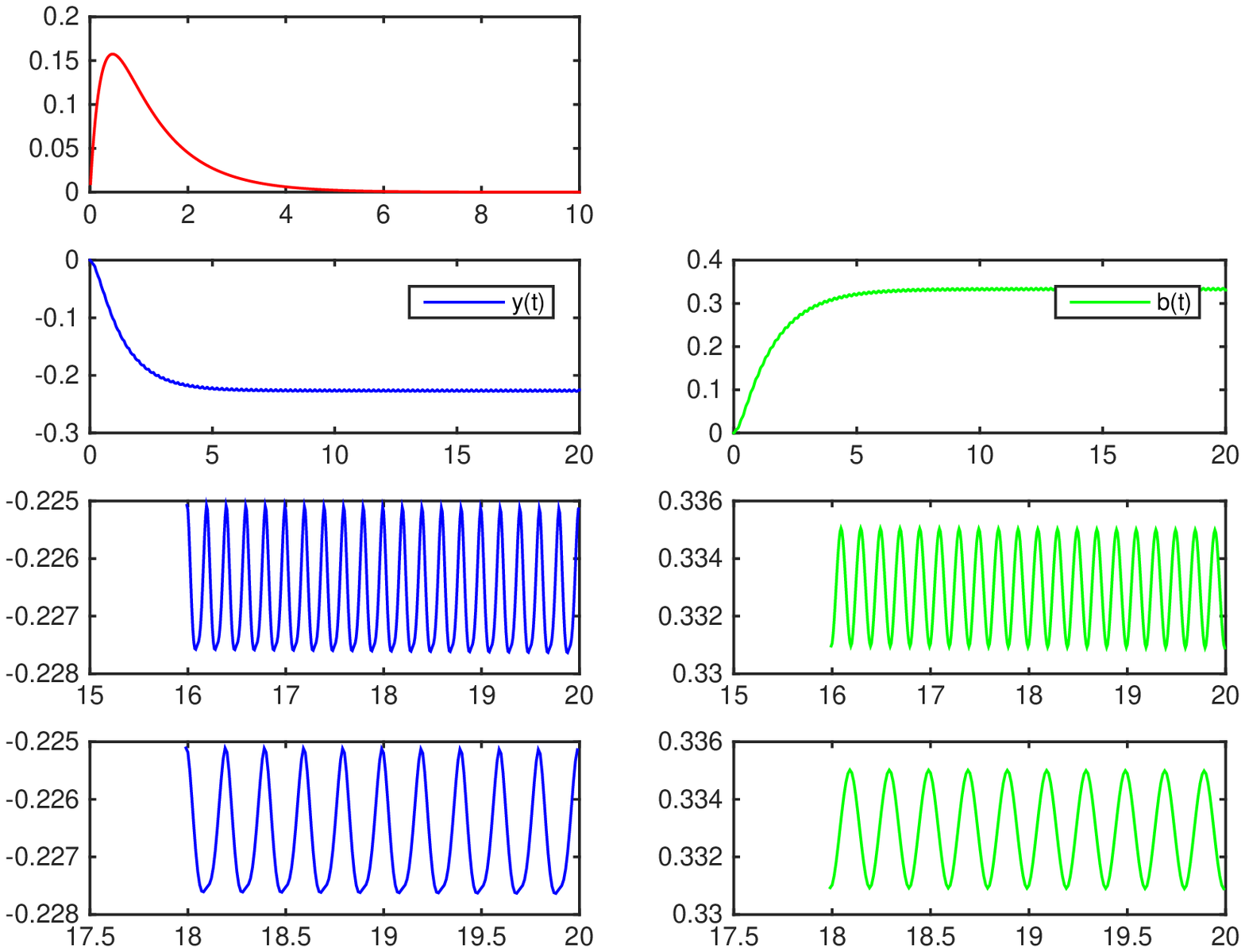}
\caption{$\g \ug 1 \, , \; \mu \ug 1 \, , \; \t \ug 5 \, , \; \a_0 \ug 10 \, , \;  \a_1 \ug 10 \, , \;   y(0)\ug y'(0)\ug b(0)\ug b'(0) \ug0 \, , \: \q \ug 0.01$, iterations $\ug 100$. \label{f3}}
\end{figure}
Because of this preliminary results, we drop the parameter $\g$ from now on as we fix  $\g \ug -1$.

\item{\bfseries{Parameter} $\mathbf{\q}$} 

This parameters represent the time-sampling step of the system. Under our practical assumptions also the time-spacing of the examples. It represent a crucial parameters of the system and we will study it after the second order operator. For now we fix $\q \ug 0.01$
\item{\bfseries{Initial Conditions}}

Because of the asymptotic behavior of the solution of (\ref{diff}), it is reasonable to assume that different Initial Conditions do not produce relevant changes in the weights at the end of optimization. Indeed in Fig.\ref{f6} and Fig.\ref{f8} we can see that the final values of the weights are in the same range of Fig.\ref{f1}, but the initial oscillation is different, due to the different starting points and derivatives. We can also see that for very different initial conditions, the time that the algorithm spent on reach the steady state is almost the same, maybe because the terms $\z_h$, which reflect the gradient, balance these differences. As in the previous case, from now on we drop the specification of initial conditions assuming them null.

\begin{figure}[H]
\hspace{-0.8cm}
\includegraphics[scale=0.7]{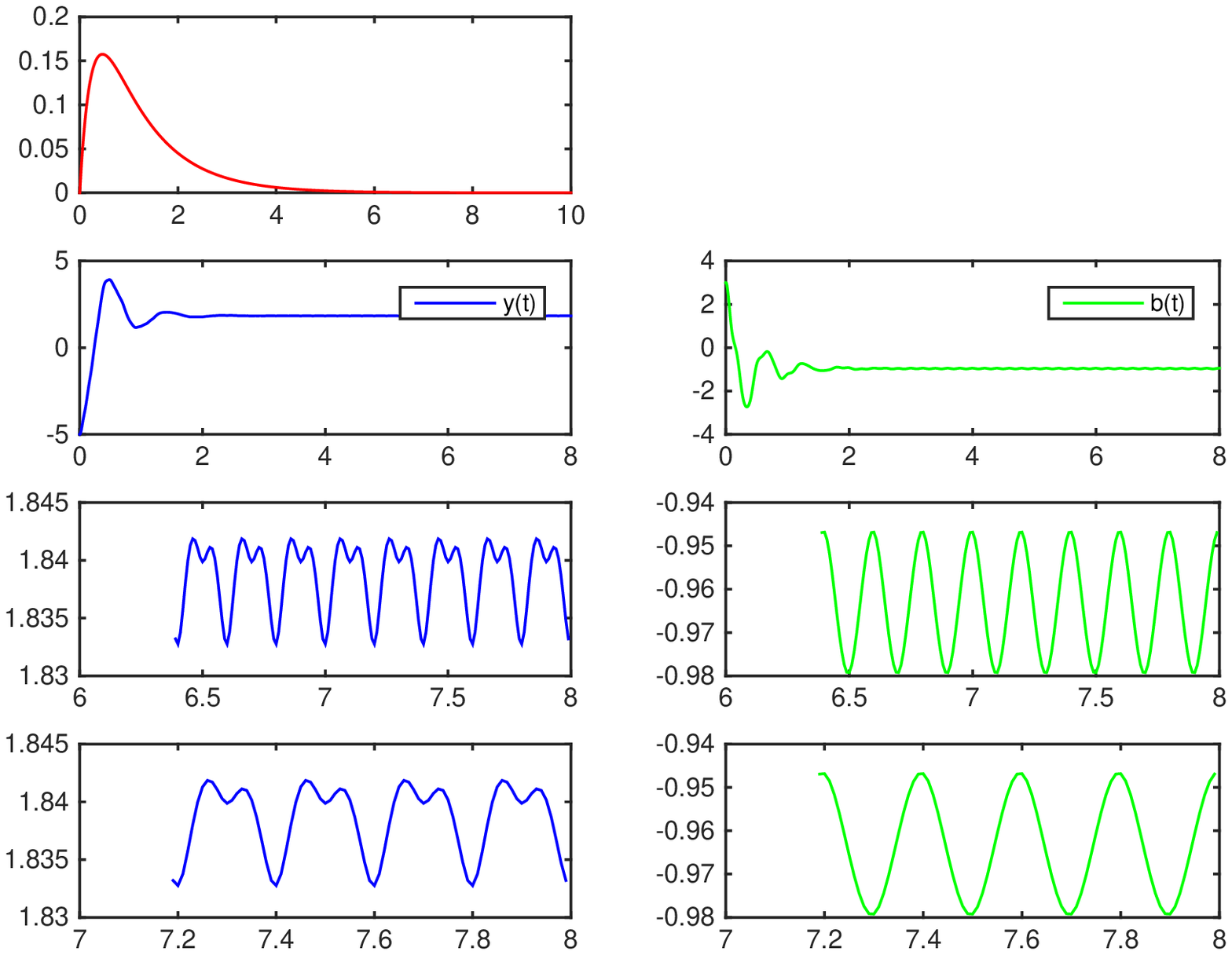}
\caption{$\mu \ug 1 \, , \; \t \ug 5 \, , \; \a_0 \ug 1 \, , \;  \a_1 \ug 1 \, , \;   y(0)\ug -5  \, , \;  y'(0)\ug -2 \, , \;  b(0)\ug 3 \, , \;  b'(0) \ug 4$, iterations $\ug40$. \label{f6}}
\end{figure}

%\begin{figure}[H]
%\hspace{-0.8cm}
%\includegraphics[scale=0.7]{./figura7.eps}
%\caption{$\mu \ug 1 \, , \; \t \ug 5 \, , \; \a_0 \ug 1 \, , \;  \a_1 \ug 1 \, , \;   y(0)\ug 25   \, , \;  y'(0)\ug 7  \, , \;  b(0)\ug -20  \, , \;  b'(0) \ug -5 $, iterations $\ug40$. \label{f7}}
%\end{figure}

\begin{figure}[H]
\hspace{-0.8cm}
\includegraphics[scale=0.7]{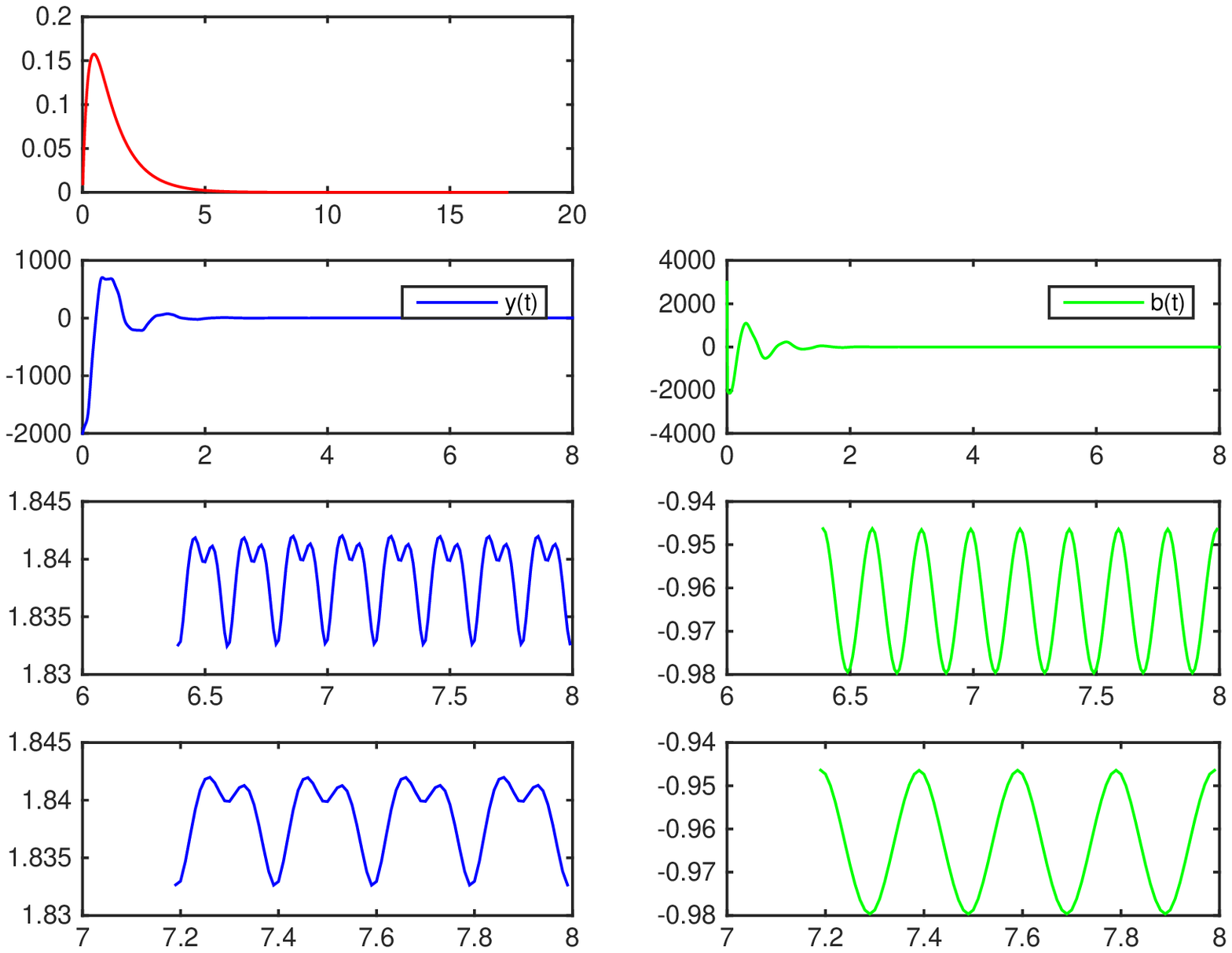}
\caption{$\mu \ug 1 \, , \; \t \ug 5 \, , \; \a_0 \ug 1 \, , \;  \a_1 \ug 1 \, , \;   y(0)\ug -2000   \, , \;  y'(0)\ug -1000 \, , \;  b(0)\ug 3000  \, , \;  b'(0) \ug 500$, iterations $\ug40$. \label{f8}}
\end{figure}

\item{\bfseries{Parameter} $\mathbf{\t}$}

This parameter come from the exponent in the dissipation term $\psi(t) \ug e^{\t t}$ and influence the solution direct in the structure of the functions $g(t)$ and $y^o(t)$ ( see section\ref{appendix}). If we decrease $\t \ug 2$, we have that $g(t) \rightarrow 0$ slowly(Fig.\ref{f9}), the initial transient phase is longer and oscillation as a greater amplitude. On the opposite,  if we set $\t \ug10$ we have the reverse effect. In the steady state, we can observe that as $\t$ increases, the average value of the weights decreases($y(t) \! \sim \! 1.956$ for $\t \ug 2$, $y(t)\! \sim \! 1.835$ for $\t \ug 5$, $y(t)\! \sim \! 1.665$ for $\t \ug10$ ), so as we can think to a regularization effect.
\begin{figure}[H]
\hspace{-0.8cm}
\includegraphics[scale=0.7]{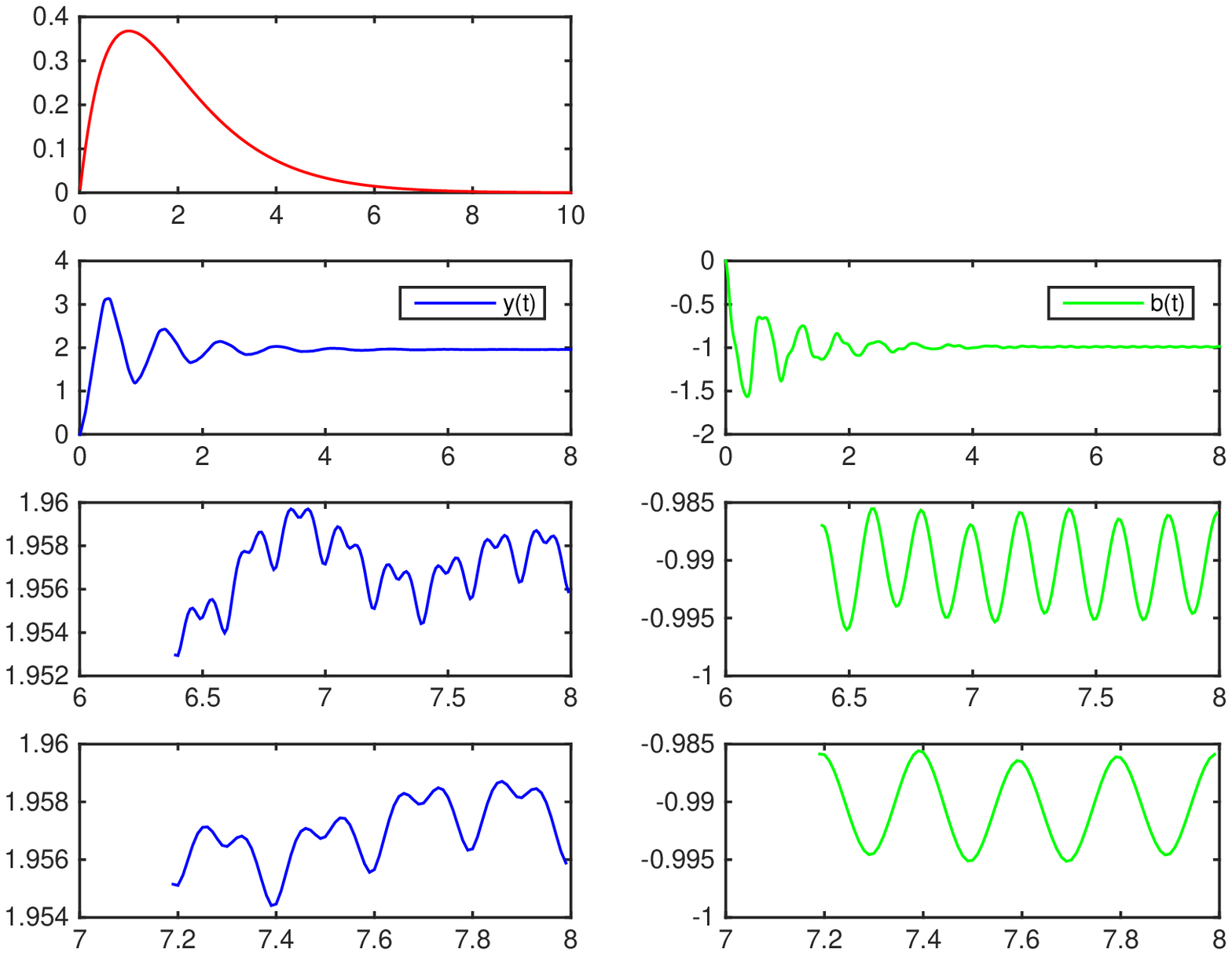}
\caption{$\mu \ug 1 \, , \; \t \ug 2 \, , \; \a_0 \ug 1 \, , \;  \a_1 \ug 1$, iterations $\ug40$. \label{f9}}
\end{figure}

\begin{figure}[H]
\hspace{-0.8cm}
\includegraphics[scale=0.7]{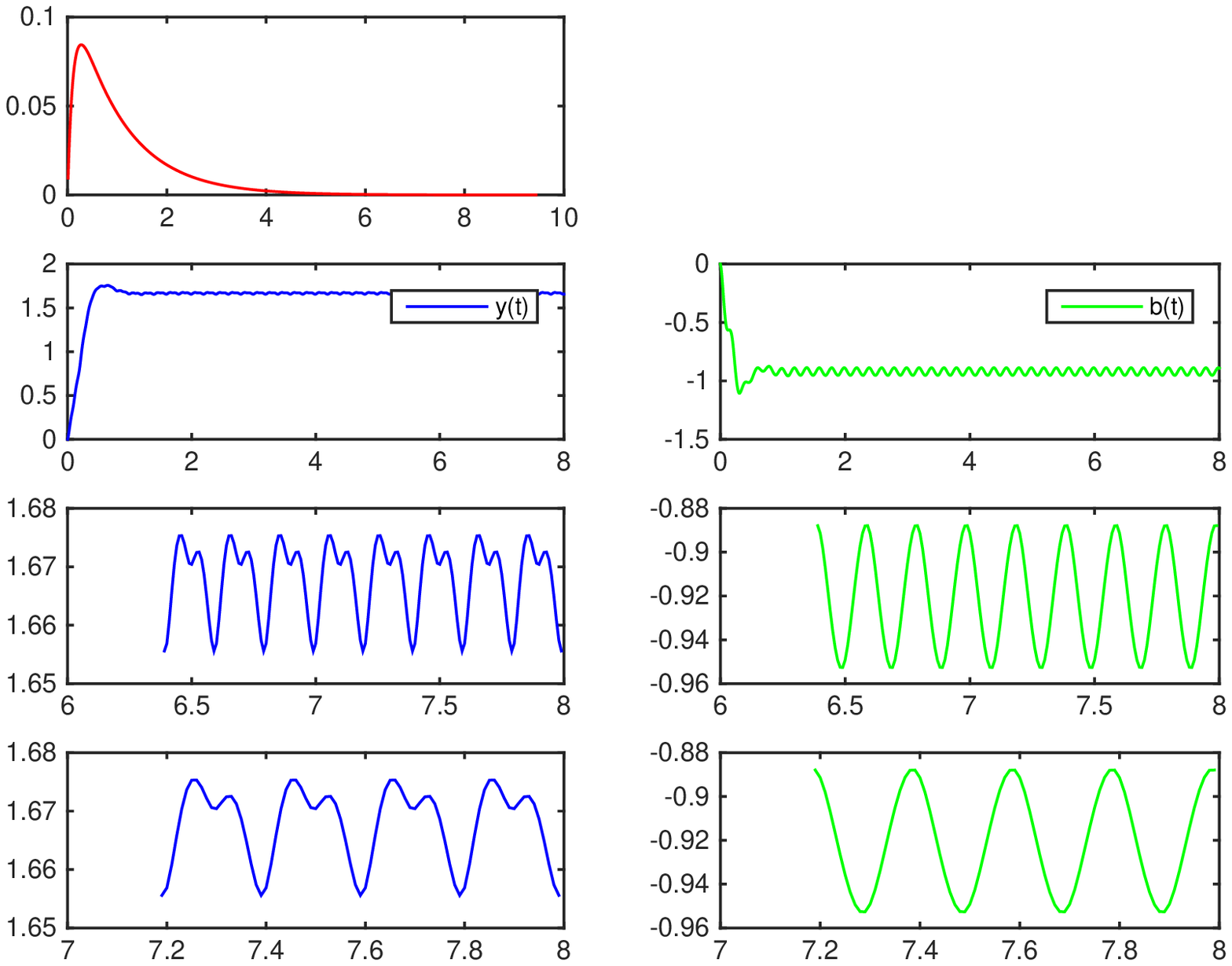}
\caption{$\mu \ug 1 \, , \; \t \ug 10 \, , \; \a_0 \ug 1 \, , \;  \a_1 \ug 1$, iterations $\ug40$. \label{f10}}
\end{figure}

\item{\bfseries{Parameter} $\mathbf{\a_0,\a_1}$}

Also these parameters affect the solution of (\ref{diff}). They are expected to act as a regularization parameters, since they appear directly in the construction of term $K$. A larger value of $\a_0$ should lead to a weights with a smaller magnitude. $\a_1$ should impose a smaller derivatives, i.e. a smooth variations of the weights during time. These hypothesis are confirmed by Fig.\ref{f11},  Fig.\ref{f12}, Fig.\ref{f13}. In Fig.\ref{f12}, we can notice that a larger $\a_1$ gives not only the expected smoothness, but also a stronger regularization effect w.r.t. $\a_0$ in Fig.\ref{f11}.

\begin{figure}[H]
\hspace{-0.8cm}
\includegraphics[scale=0.7]{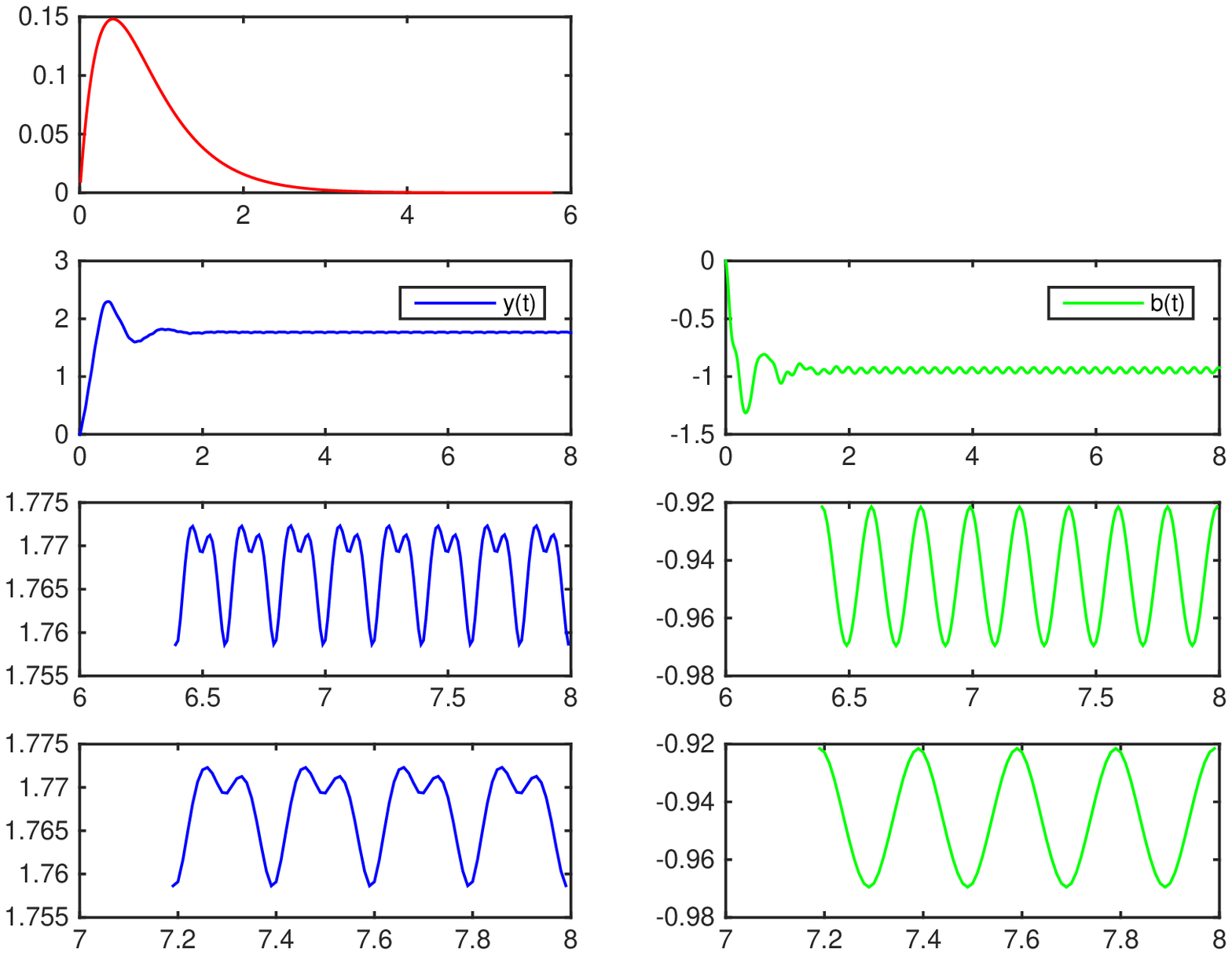}
\caption{$\mu \ug 1 \, , \; \t \ug 5 \, , \; \a_0 \ug 3 \, , \;  \a_1\ug 1$, iterations $\ug40$. \label{f11}}
\end{figure}

\begin{figure}[H]
\hspace{-0.8cm}
\includegraphics[scale=0.7]{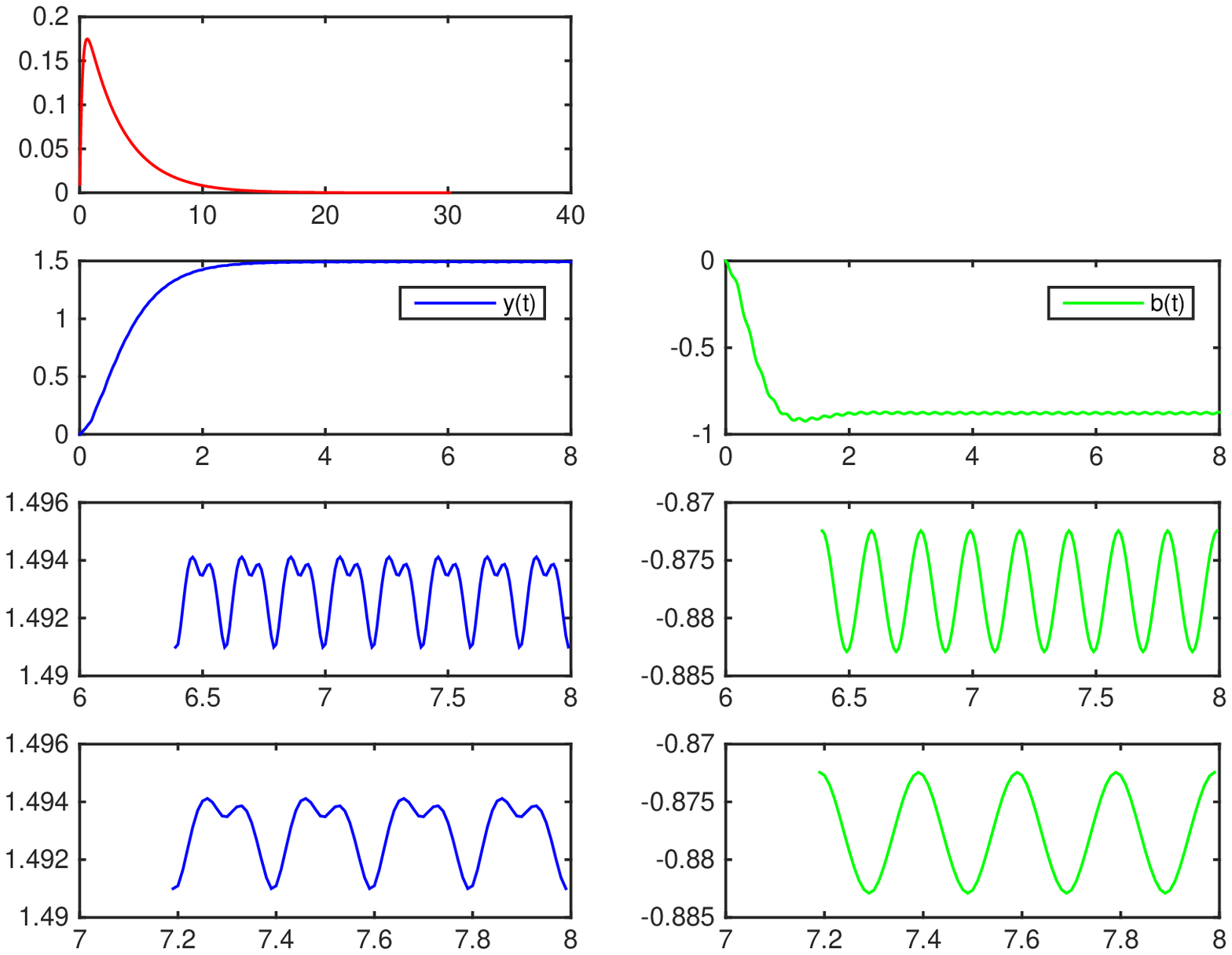}
\caption{$\mu \ug 1 \, , \; \t \ug 5 \, , \; \a_0 \ug 1 \, , \;  \a_1 \ug 3$, iterations $\ug40$. \label{f12}}
\end{figure}

\begin{figure}[H]
\hspace{-0.8cm}
\includegraphics[scale=0.7]{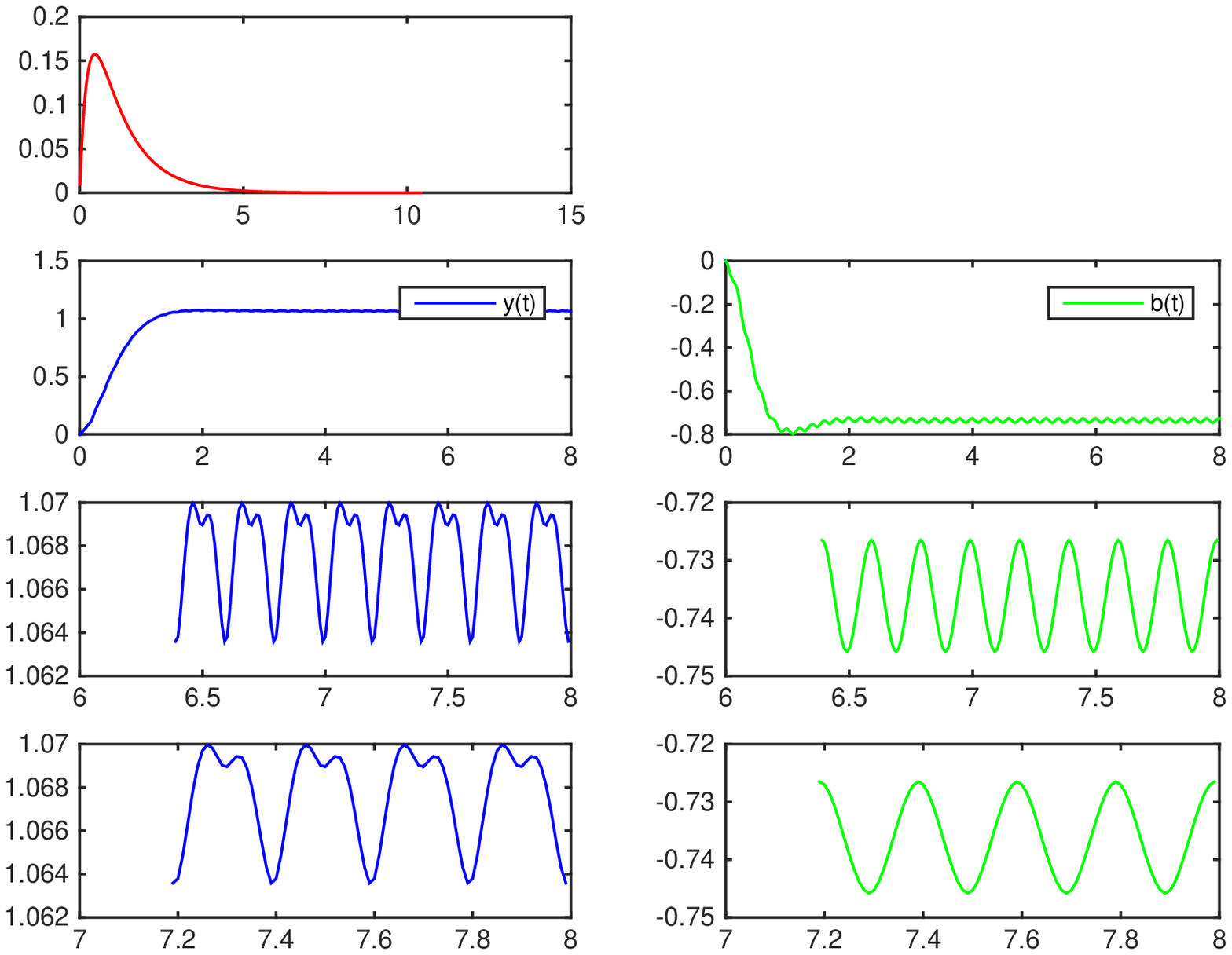}
\caption{$\mu \ug 1 \, , \; \t \ug 5 \, , \; \a_0 \ug 3 \, , \;  \a_1 \ug 3$, iterations $\ug40$. \label{f13}}
\end{figure}

\item{\bfseries{Parameter} $\mathbf{\mu}$}

$\mu$ is the correspondent of the \emph{mass} in physics, so it should represent the inertia of the weights, i.e. how we want to allow to the system to change them at each step (w.r.t. the penalty term $V$). Furthermore, we can see in (\ref{upfo}) that it can be use to represent a sort of learning rate (to follow a parallel with gradient descent again). This is confirmed in Fig.\ref{f14} and Fig.\ref{f15} where a smaller value of $\mu$ gives more importance to the optimization than to the regularization.

\begin{figure}[H]
\hspace{-0.8cm}
\includegraphics[scale=0.7]{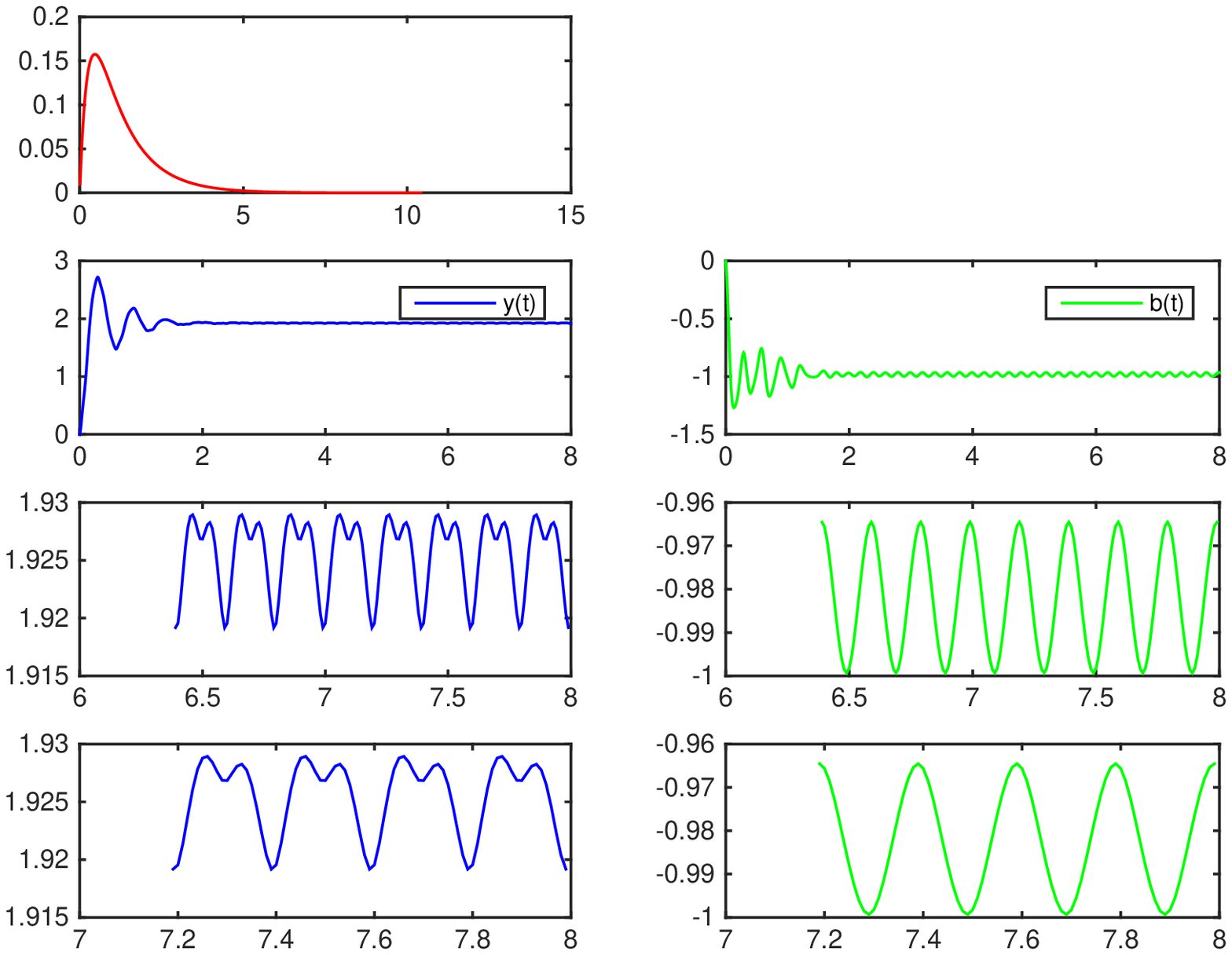}
\caption{$\mu \ug 0.5 \, , \; \t \ug 5 \, , \; \a_0 \ug 1 \, , \;  \a_1 \ug 1$, iterations $\ug40$. \label{f14}}
\end{figure}

\begin{figure}[H]
\hspace{-0.8cm}
\includegraphics[scale=0.7]{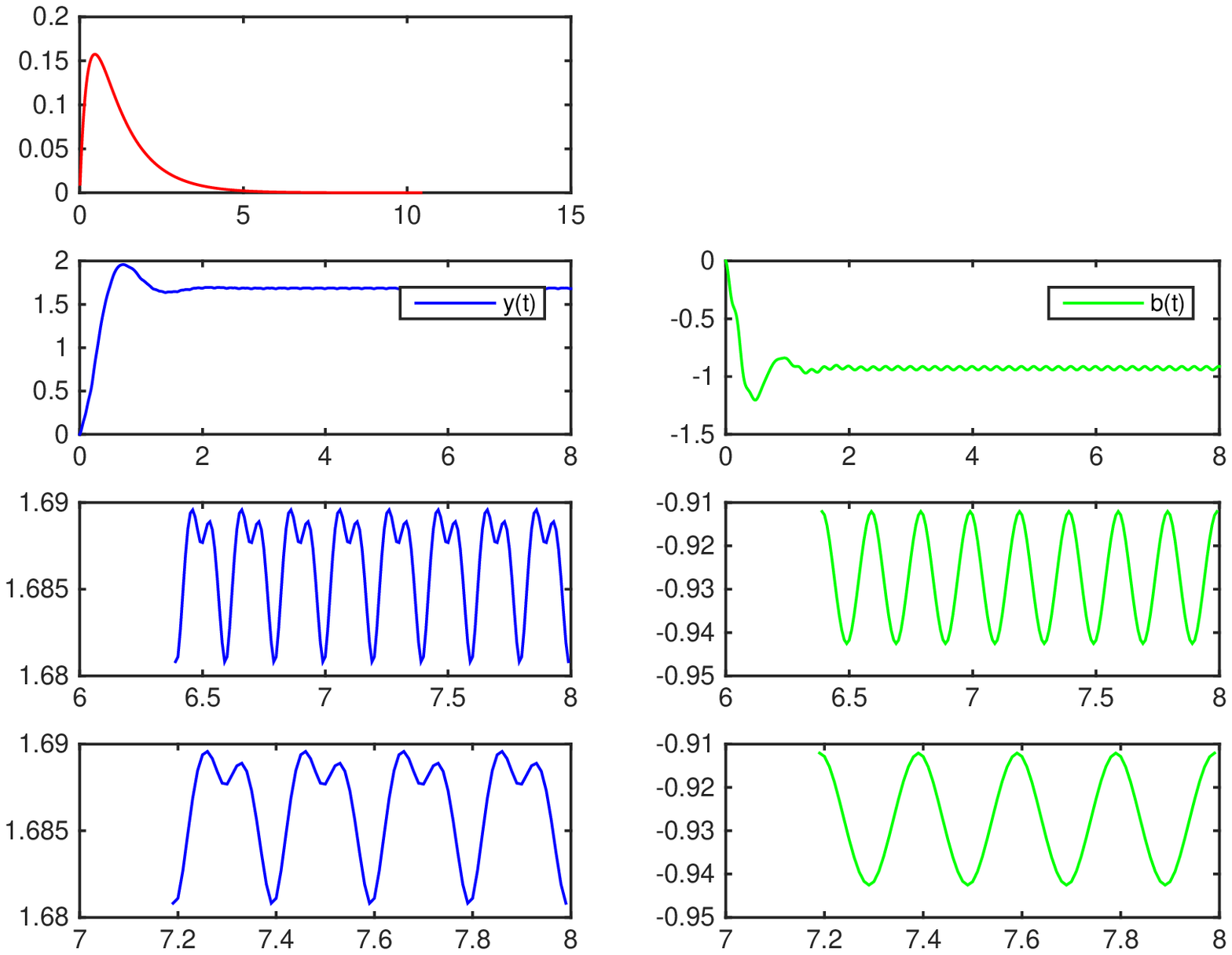}
\caption{$\mu \ug 2 \, , \; \t \ug 5 \, , \; \a_0 \ug 1 \, , \;  \a_1 \ug 1$, iterations $\ug40$. \label{f15}}
\end{figure}

If we try to grow up with $\mu$, we find that this increment maintain this behavior (unbalancing towards  regularization). On the other hand, if we choose a $\mu  \leq 0.4$, the system diverges(Fig.\ref{f15b}). 

\end{description}

\begin{figure}[H]
\hspace{-0.8cm}
\includegraphics[scale=0.7]{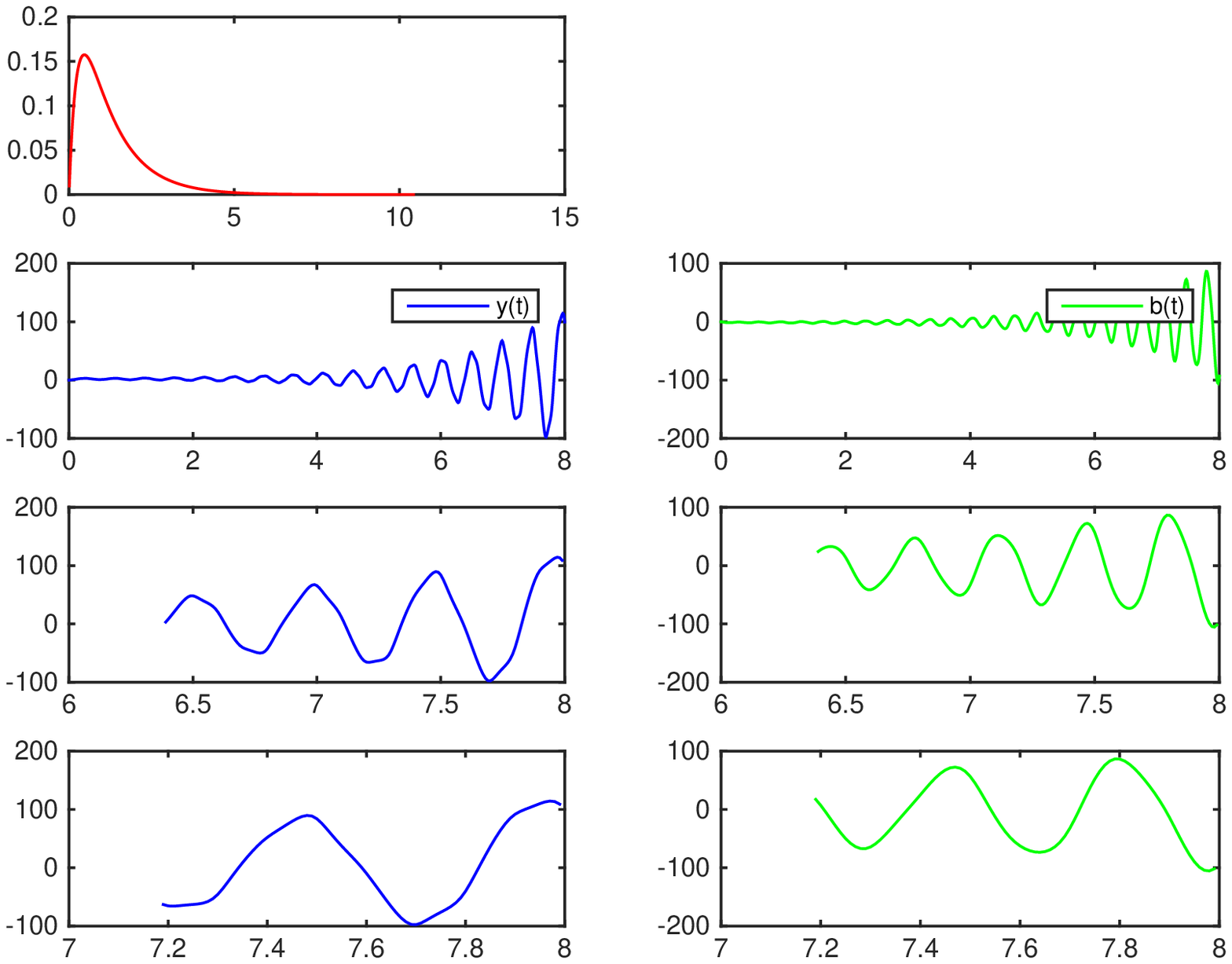}
\caption{$\mu \ug 0.4 \, , \; \t \ug 5 \, , \; \a_0 \ug 1 \, , \;  \a_1 \ug 1$, iterations $\ug40$. \label{f15b}}
\end{figure}

\subsubsection{Choice of solutions}

\begin{figure}[H]
\centering
\hspace{-0.8cm}
\includegraphics[scale=0.5]{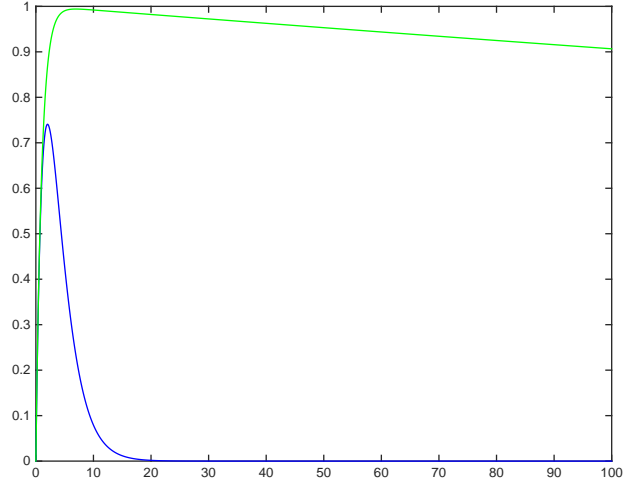}
\caption{Plot of two different Impulsive Response, the green plot represent $g$ when $\l_1 \ug -0.999 \, , \; \l_2 \ug -0.001$ and the blue one  $\l_1 \ug -0.6 \, , \; \l_2 \ug -0.4$ \label{IRexfo}}
\end{figure}

All this parameter contribute to model our system, but we have to choose how. Since we would like that the system promptly reacts to the stimuli, we want an Impulsive Response that reach its maximum as fast as possible. Moreover, its useful that the system is width enough to see all the examples more times before to forget them. Once we have chosen a suitable $\t$ (w.r.t. the Training Set, see Section \ref{taustudy}), we can model the parameters $\a_j$ so as to satisfy these request. Often is more convenient to choose the directly the solutions to build our system (see Section \ref{findparam}). In Fig.\ref{IRexfo} we can see that if we chose a solution close to $0$, and then the other close to $\t$ (since $\t \ug - (\l_1 + \l_2)$) we have a longer $g$ (green plot), whereas we go in the other direction if the solutions are similar (blue plot).

% % % % \input ./Subsections/SecondOrderOperator

\subsection{Second Order Operator}\label{2ndOrd}% $T \ug \alpha_0 \! + \! \alpha_1 D \! + \! \alpha_2 D^2$}

Under the practical assumptions of the previous section, we study the implementation of the case in which the term $K$ is composed by the second order linear differential operator $T \ug \alpha_0 \! + \! \alpha_1 D \! + \! \alpha_2 D^2$. The \emph{Eulero-Lagrange} equation lead this time to a fourth order linear differential equation:

\begin{equation}\label{diff4}
D^4y + \b_3 D^3 y + \b_2 D^2 y + \b_1 D y + \b_0  y +  \frac{\gamma}{\mu \alpha_2^2 } \sum_{k =1}^l (u_k y_k + b_k - \by_k)u_k \cdot \delta (t-t_k) =0 
\end{equation}
where:
 \begin{equation}\label{coeff4}
 \begin{array}{rcl}
 \b_0 \! & \ug & \frac{\a_0 \a_2 \t^2 - \a_0 \a_1 \t  + \a_0 ^2}{a_2^2}  \\
  \b_1 \! & \ug & \frac{\a_1 \a_2 \t^2 + (2 \a_0 \a_2 - \a_1^2)\t}{\a_2^2} \\
 \b_2 \! & \ug & \frac{\a_2^2 \t^2 + \a_1 \a_2 \t + 2 \a_0 \a_2 - \a_1^2}{\a_2^2}\\
 \b_3 \! & \ug & 2\t 
 \end{array}
 \end{equation}
  This time the \emph{Routh-Hurwitz conditions} requires:

 \begin{equation}\label{RH4}
 \begin{array}{rcll}
\b_i                       &  >    & & 0 \; , \quad i=0,...,3 \\
\b_3 \b_2              & >     &\b_1 & \\
\b_3 \b_2 \b_1     & >      &\b_1^2 + \b_3^2 \b_0 & 
 \end{array}
 \end{equation}
 The updating formula is then\footnote{again see section \ref{appendix} for details about practical calculation}:
\begin{equation}\label{fbp2}
y(t)=y^o(t) - \; \frac{\g}{\mu \, \a_2^2 } \cdot \sum_{k}^l \z_h \cdot g(t-t_k)
\end{equation} 
where we can notice a sign flip before the second term w.r.t. (\ref{upfo}) due to the even order of $T$. This make us expect (because of the parallel with gradient descent again) that this time our system is stable for an opposite sign of $\g$ w.r.t. the first order operator studied in section \ref{fooe}. In practice, we can observe a more complicated behavior due to the kind of the solutions of (\ref{diff4}).

\subsubsection{Experimental results}

This time we start by observing that there are different possibilities in the kind of the solutions of (\ref{diff4}): 

\begin{description}
\item{{\bfseries(1)Four distinct real solution}}

We start with the case $\t \ug 4 \, , \; \a_0 \ug 0.8 \, , \;  \a_1 \ug 1.6 \, , \;  \a_2 \ug 0.8 $. As we can see in Fig.\ref{fs16} the system is divergent for $\g \ug - 1$. Also in the case $\g \ug 1$ in Fig.\ref{fs16b} we have divergence, but we can notice a different oscillation of the weights. We can see in (\ref{fbp2}) that the second term is multiplied by the factor $\g / ( \mu \, \a_2^2 )$. Since $\a_2 <1$ we can apply some of the considerations done in Section \ref{fooe} about the parameter $\mu$. This divergence is maybe due to a too higher balancing on the gradients term, indeed if we set $\mu \ug 4$ we have convergence to the desired values Fig. \ref{fs17}. We can obtain convergence by increasing $\mu$ also when $\g \ug -1$ Fig. \ref{fs17b} but with futile values of the weights.

\begin{figure}[H]
\hspace{-0.8cm}
\includegraphics[scale=0.7]{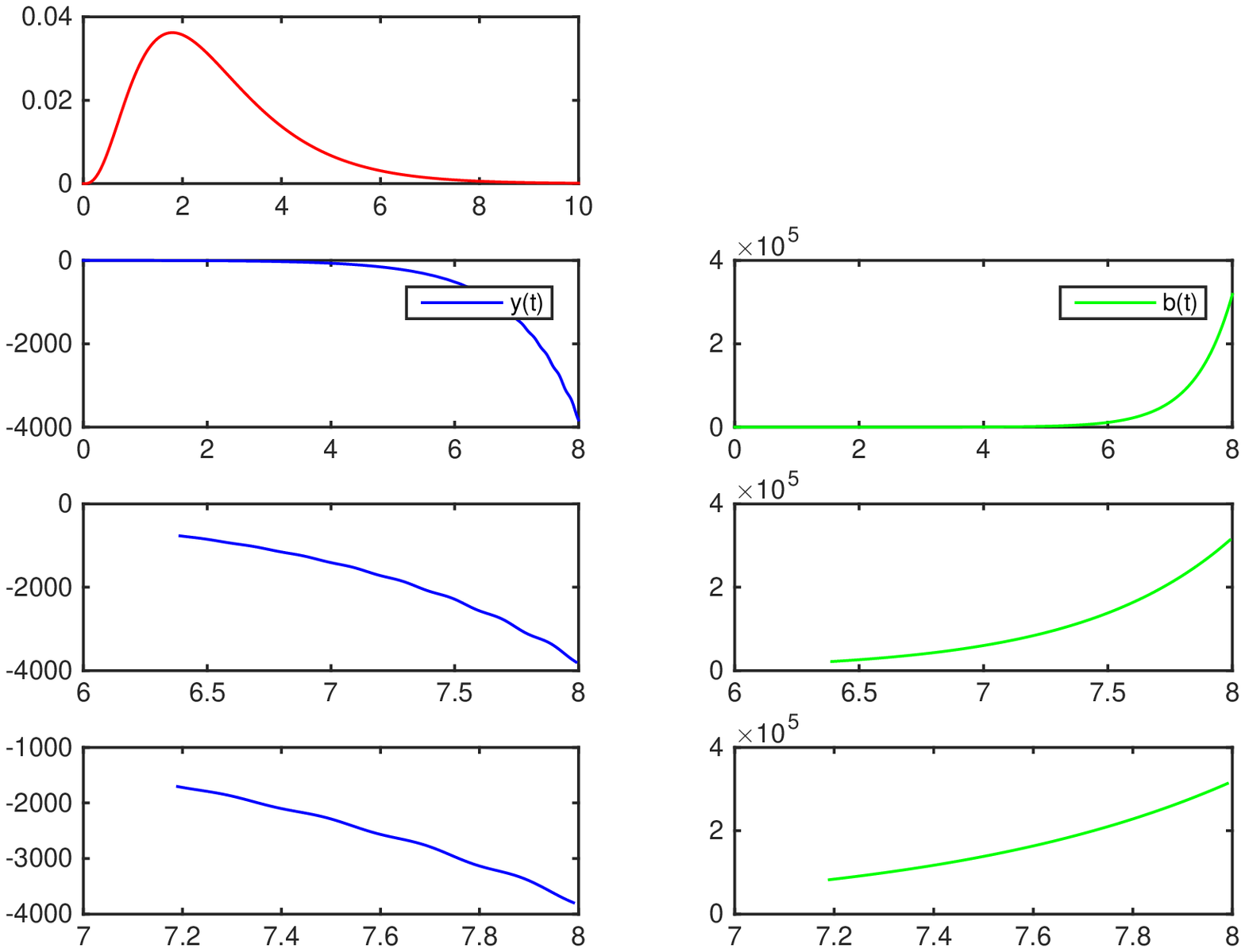}
\caption{$\g \ug -1 \, , \; \mu \ug 1 \, , \; \t \ug 4 \, , \; \a_0 \ug 0.8 \, , \;  \a_1 \ug 1.6 \, , \;  \a_2 \ug 0.8 \, , \;  \q \ug 0.01$, null Initial Conditions , iterations $\ug40$. \label{fs16}}
\end{figure}
 
 \begin{figure}[H]
\hspace{-0.8cm}
\includegraphics[scale=0.7]{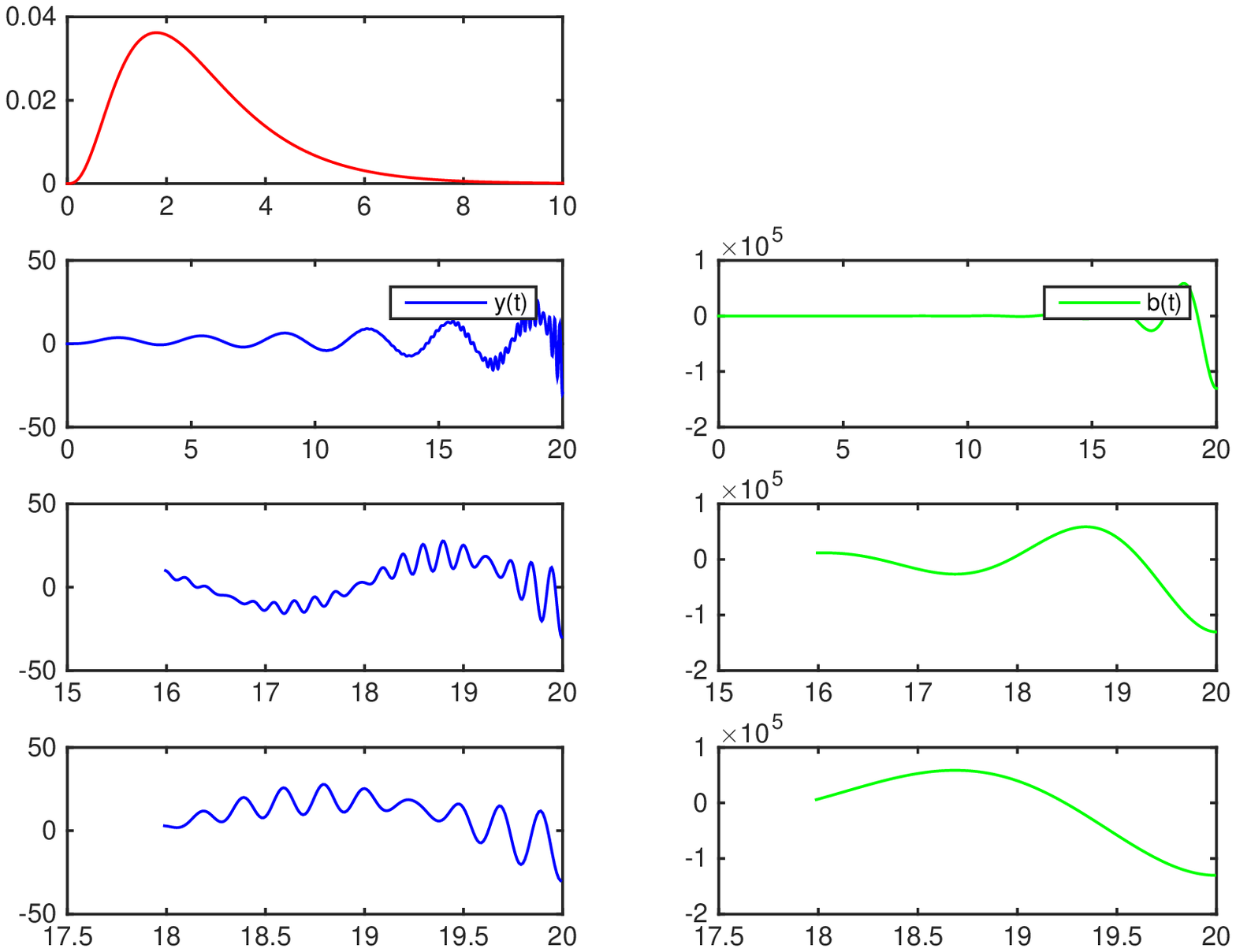}
\caption{$\g \ug 1 \, , \; \mu \ug 1 \, , \; \t \ug 4 \, , \; \a_0 \ug 0.8 \, , \;  \a_1 \ug 1.6 \, , \;  \a_2 \ug 0.8 \, , \;  \q \ug 0.01$, null Initial Conditions , iterations $\ug100$. \label{fs16b}}
\end{figure}

 \begin{figure}[H]
\hspace{-0.8cm}
\includegraphics[scale=0.7]{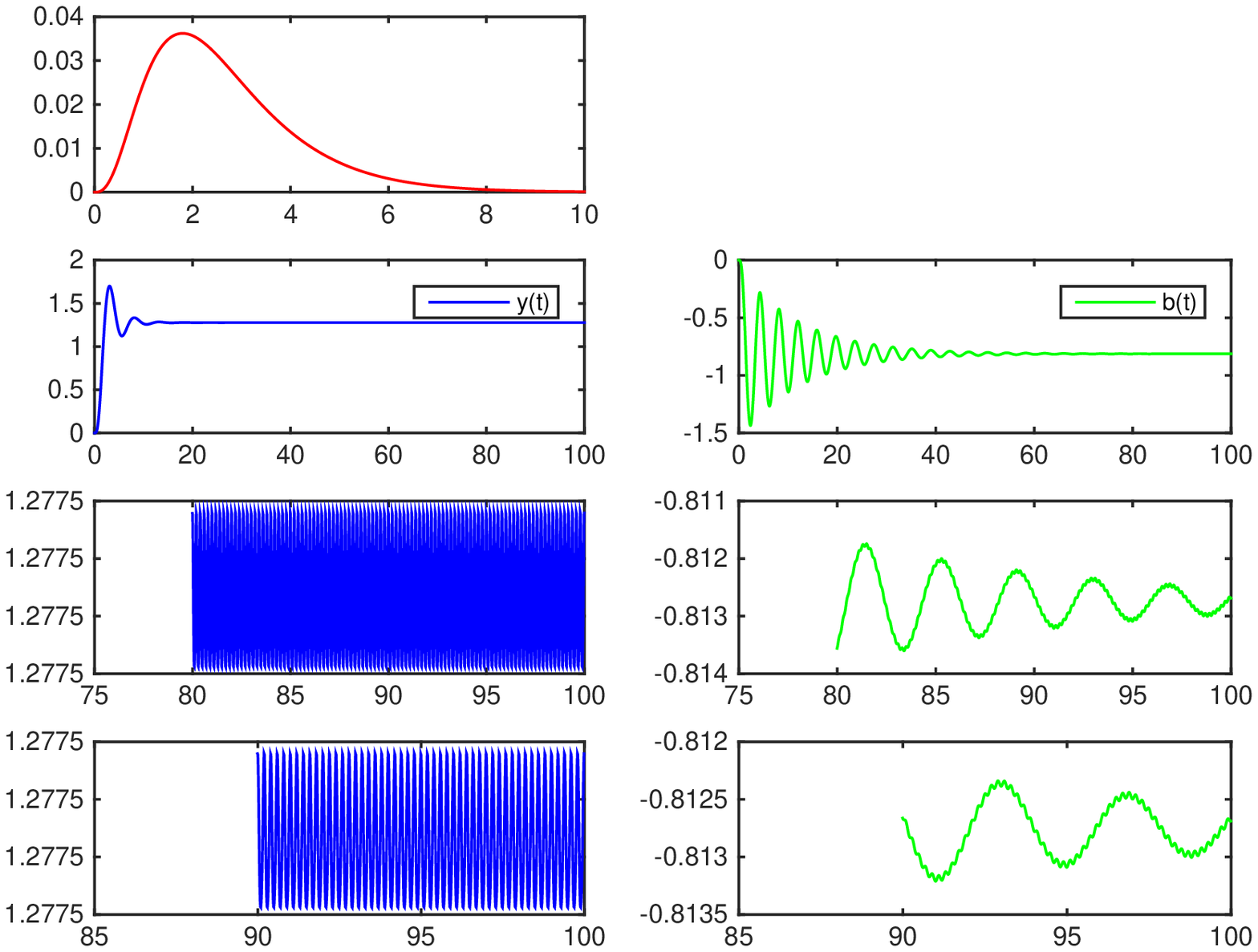}
\caption{$\g \ug 1 \, , \; \mu \ug 4 \, , \; \t \ug 4 \, , \; \a_0 \ug 0.8 \, , \;  \a_1 \ug 1.6 \, , \;  \a_2 \ug 0.8 \, , \;  \q \ug 0.01$, null Initial Conditions , iterations $\ug500$. \label{fs17}}
\end{figure}

 \begin{figure}[H]
\hspace{-0.8cm}
\includegraphics[scale=0.7]{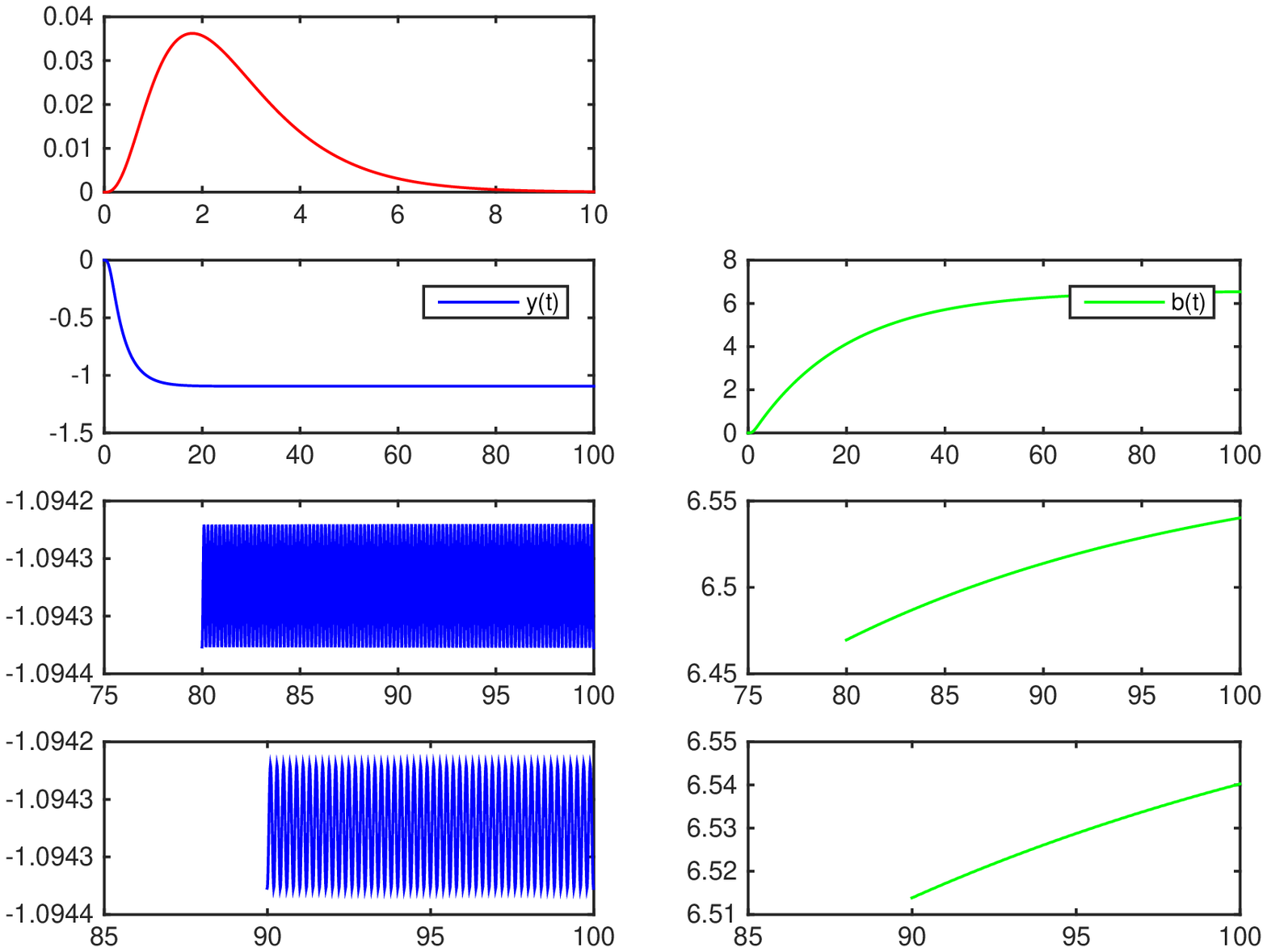}
\caption{$\g \ug -1 \, , \; \mu \ug 20 \, , \; \t \ug 4 \, , \; \a_0 \ug 0.8 \, , \;  \a_1 \ug 1.6 \, , \;  \a_2 \ug 0.8 \, , \;  \q \ug 0.01$, null Initial Conditions , iterations $\ug500$. \label{fs17b}}
\end{figure}

We have this kind of solution also for  $ 2.2 \leq \t \leq 4$ and for $ 6.6 \leq \t \leq 7.2$, with the same value of $\a_j$. When $\t \ug 2.2$ we need a bigger value of $\mu$ to achieve convergence Fig. \ref{fs18}. This is because the impulse response has a bigger maximum than before, and then also the coefficients which multiply the gradients are bigger, so that we need a bigger balancing $\mu$.

\begin{figure}[H]
\hspace{-0.8cm}
\includegraphics[scale=0.7]{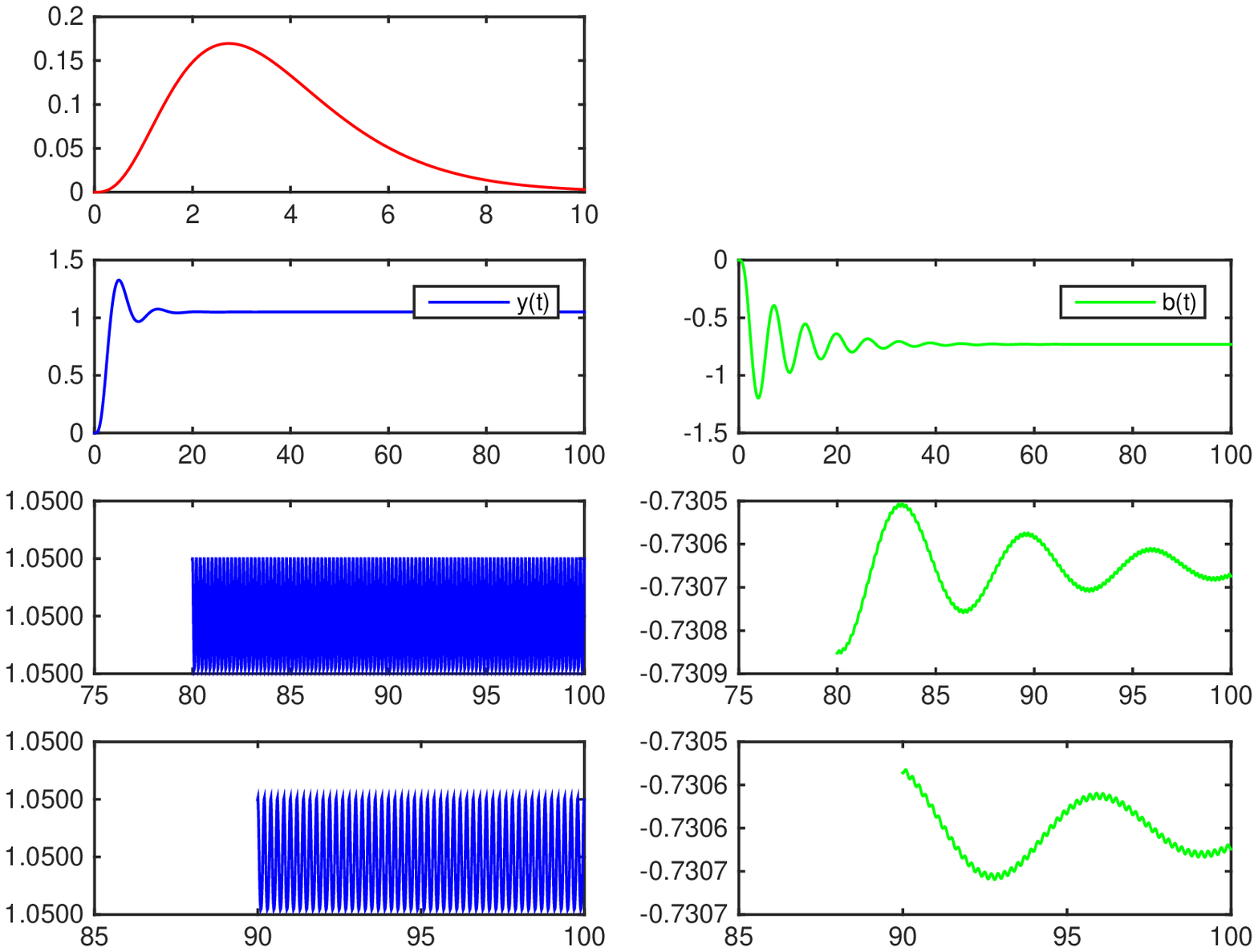}
\caption{$\g \ug 1 \, , \; \mu \ug 40 \, , \; \t \ug 2.2 \, , \; \a_0 \ug 0.8 \, , \;  \a_1 \ug 1.6 \, , \;  \a_2 \ug 0.8 \, , \;  \q \ug 0.01$, null Initial Conditions , iterations $\ug500$. \label{fs18}}
\end{figure}

When $\t \ug 6.8$ these first conclusion are confirmed in Fig.\ref{fs19}, where $\mu \ug 1$ is enough for convergence. This is because  the solutions to (\ref{diff4}) are different, but we can comparing again the impulse response that assume a smaller value than before.

 \begin{figure}[H]
\hspace{-0.8cm}
\includegraphics[scale=0.7]{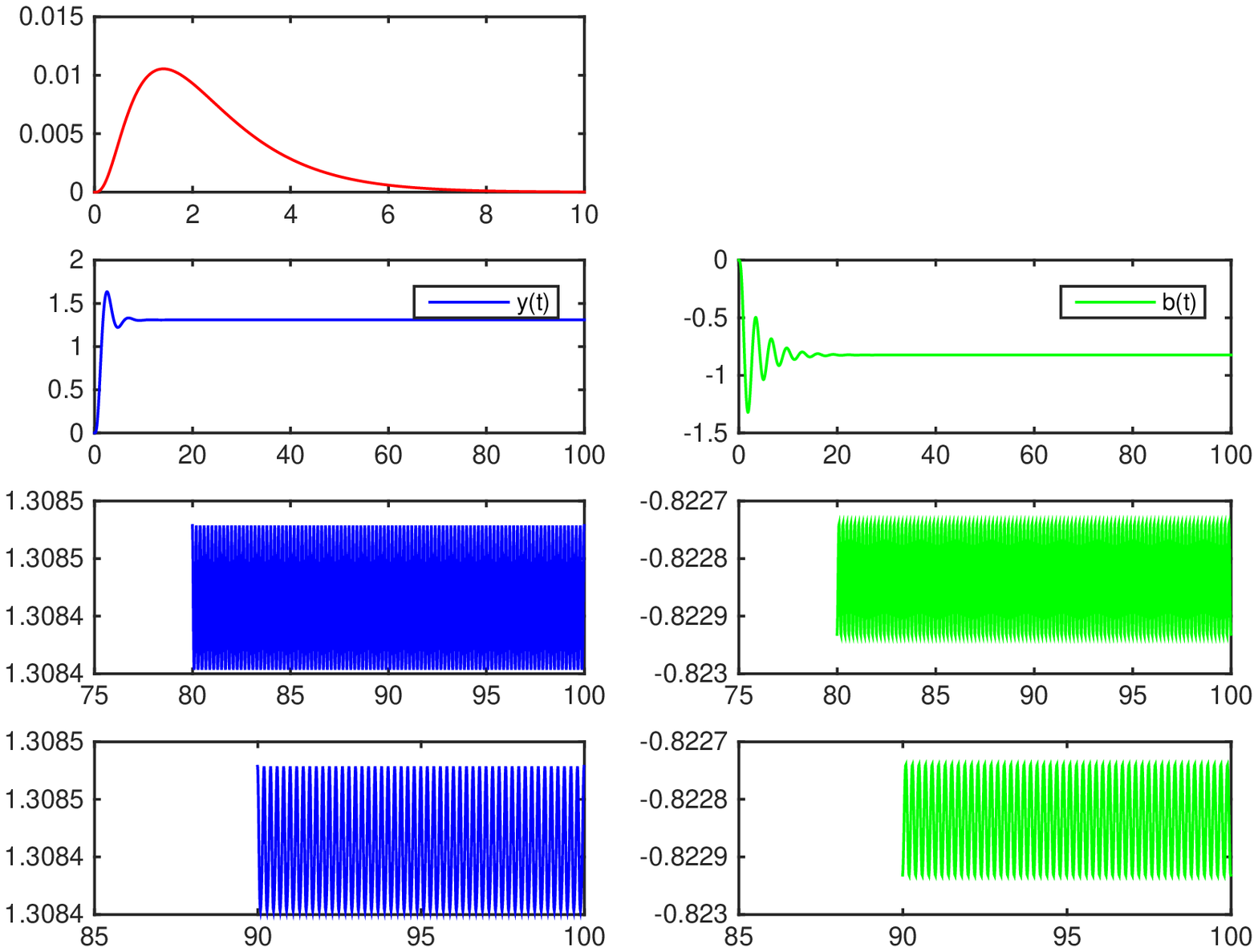}
\caption{$\g \ug 1 \, , \; \mu \ug 1 \, , \;  \t \ug 6.8 \, , \; \a_0 \ug 0.8 \, , \;  \a_1 \ug 1.6 \, , \;  \a_2 \ug 0.8 \, , \;  \q \ug 0.01$, null Initial Conditions , iterations $\ug500$. \label{fs19}}
\end{figure}

\item{{\bfseries(2) Four real solution, 1 with multiplicity 2}}

Also in this case the system diverge when $\g \ug -1$. When $\g  \ug 1$ we report the case in Fig.\ref{so2_1} where we can see a behavior similar to case {\bfseries (1)}.

\begin{figure}[H]
\hspace{-0.8cm}
\includegraphics[scale=0.7]{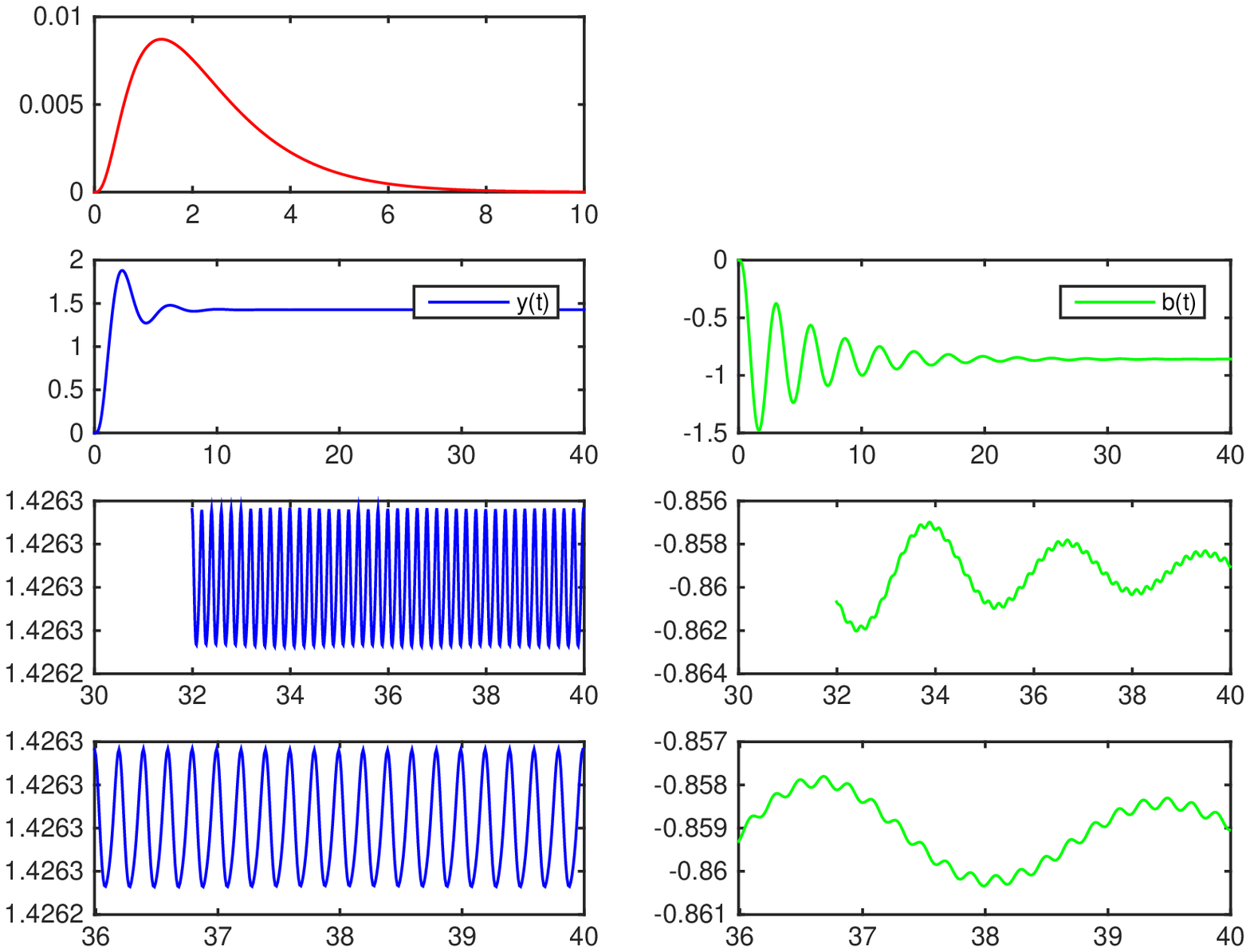}
\caption{$\g \ug 1 \, , \; \mu \ug 0.1 \, , \; \t \ug 7.4 \, , \; \a_0 \ug 0.2 \, , \;  \a_1 \ug 0.4 \, , \;  \a_2 \ug 0.2 \, , \;  \q \ug 0.01$, null Initial Conditions , iterations $\ug200$. \label{so2_1}}
\end{figure}

\item{{\bfseries(3)Two distinct real solutions, two conjugate complex solution}}

We can obtain this solutions for example for these settings of the parameters:
\begin{footnotesize}
\begin{center}
\begin{tabular}{|c|c|c|c|}
\hline
 $\mathbf{\t}$ & $\mathbf{\a_0}$ &  $\mathbf{\a_1}$ & $\mathbf{\a_2}$ \\
\hline
\hline
   5.75   & 1 & 2 & 1 \\
\hline
     7   & 1 & 2 & 1 \\
\hline
    1 & 1  & 3 & 2.25 \\
\hline
 2.25 & 1  & 3 & 2.25 \\
\hline
\end{tabular}
\end{center}
\end{footnotesize}

When we have the complex solution with real part smaller than the real solutions , part of the Impulse Response without oscillation (the one coming from the real solutions) disappears before the other one, and if the imaginary part is big enough we have an oscillation that lead to a more complex behavior and then to instability. We can choose (see Section \ref{findparam}) $\l_{1,2} \ug -0.1 \pm i \, , \; \l_3 \ug -1.2\, , \; \l_4 \ug -1$ and we have the situation in Fig.\ref{comp1}, where a big regularization is required to convergence.

\begin{figure}[H]
\hspace{-0.8cm}
\includegraphics[scale=0.7]{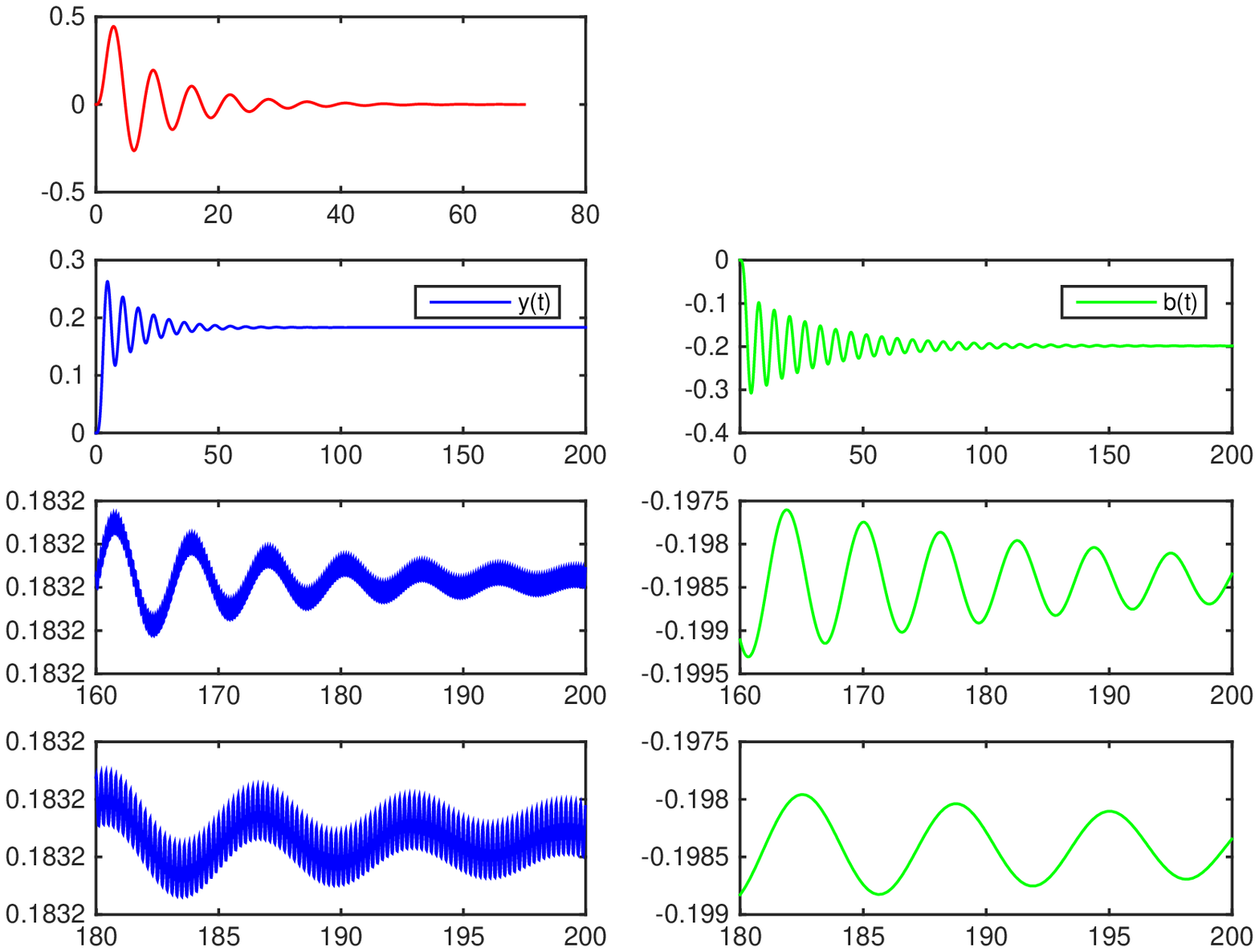}
\caption{$\g \ug 1 \, , \; \mu \ug 3000  \, , \; $ Solutions: $\l_{1,2} \ug -0.1 \pm i \, , \; \l_3 \ug -1.2\, , \; \l_4 \ug -1  \, , \;  \q \ug 0.01$, null Initial Conditions , iterations $\ug1000$. \label{comp1} }
\end{figure}

\item{{\bfseries(4)One real solution with multiplicity 2 , two conjugate complex solution}}

We can obtain this solutions for example when:
\begin{footnotesize}
\begin{center}
\begin{tabular}{|c|c|c|c|}
\hline
 $\mathbf{\t}$ & $\mathbf{\a_0}$ &  $\mathbf{\a_1}$ & $\mathbf{\a_2}$ \\
\hline
\hline
   8.2   & 0.2 & 1.2 & 1.8 \\
\hline
18.4 & 0.2 & 1.2 & 1.8 \\
\hline
\end{tabular}
\end{center}
\end{footnotesize}

and we have an analogous situation of {\bfseries (3)}.

\item{{\bfseries(5)Four complex solution, two conjugate pairs}}

We have both the case in which we have two complex conjugate pairs and the case with a complex conjugate pair with multiplicity two. They have a similar behavior depending on the magnitude of the real and imaginary parts. The real parts influence the memory of the system, whereas the imaginary parts the frequency of the sinusoidal oscillation. In each case is possible to find a value of $\mu$ which allow convergence, but often to values near to $0$ with an high-oscillatory trend similar to the one reported in Fig.\ref{comp1}, since it is due to the sinusoidal nature of the solution.

\end{description}

\begin{figure}[H]
\hspace{-0.8cm}
\includegraphics[scale=0.7]{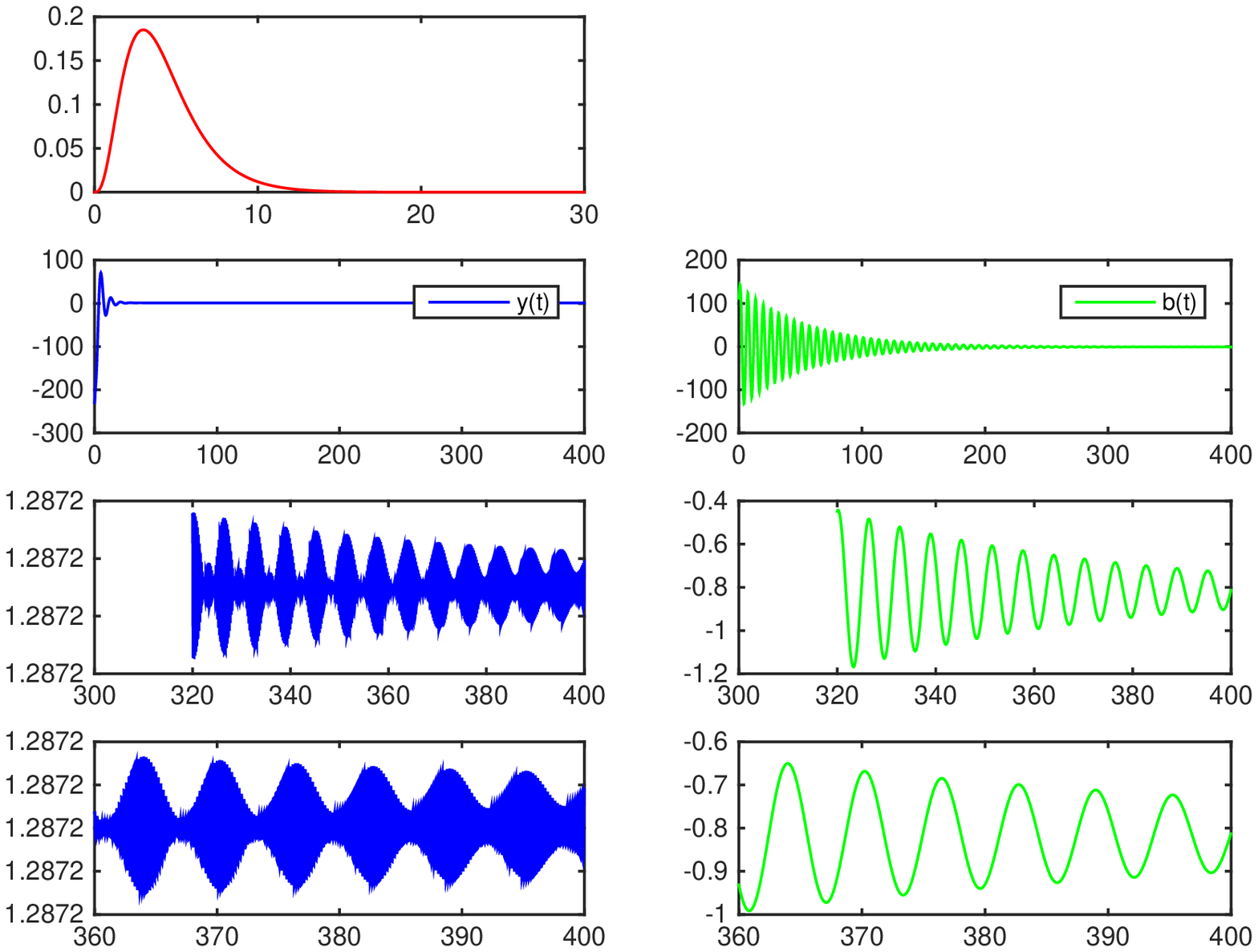}
\caption{$\g \ug 1 \, , \; \mu \ug 4 \, , \; \t \ug 2.25 \, , \; \a_0 \ug 1 \, , \;  \a_1 \ug 3 \, , \;  \a_2 \ug 2.25 \, , \;  \q \ug 0.01 \, , \;  y_0 \ug [-230 \, , \;54\, , \; 0 \, , \;-342] \, , \; b_0\ug [112 \, , \;41\, , \; 19 \, , \;0]$, iterations $\ug2000$. \label{fs4}}
\end{figure}

From this study on the solutions is clear that the parameter $\mu$ decides again the balancing between regularization and fitting, as observed for the first order operator. The behavior w.r.t. Initial Conditions is again the same as we can see in Fig. \ref{fs4}. Since we have a fourth-order differential equation we need to know the values of the first $n-1 \ug 3$ derivatives of the solution $y^o(t)$ in $t\ug 0$. We indicate I.C. with the vectors $y_0 \ug [y^o(0) \, , \; {y^o}^{(1)}(0) \, , \; {y^o}^{(2)}(0) \, , \; {y^o}^{(3)}(0)]$ and  $b_0\ug [b^o(0) \, , \; {b^o}^{(1)}(0) \, , \; {b^o}^{(2)}(0) \, , \; {b^o}^{(3)}(0)]$.

\begin{figure}[H]
\hspace{-0.8cm}
\includegraphics[scale=0.7]{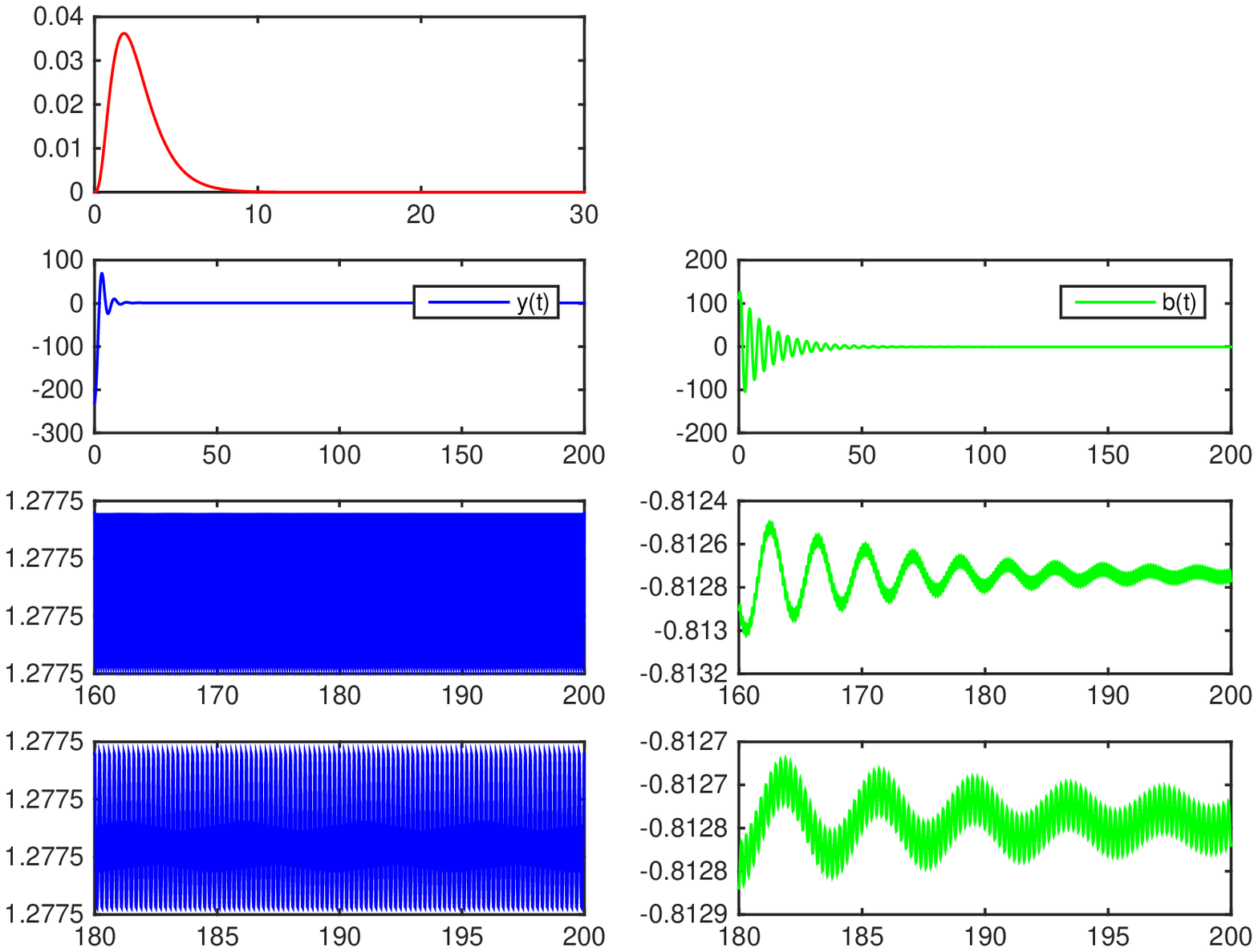}
\caption{$\g \ug 1 \, , \; \mu \ug 4 \, , \; \t \ug 4 \, , \; \a_0 \ug 0.8 \, , \;  \a_1 \ug 1.6 \, , \;  \a_2 \ug 0.8 \, , \;   \q \ug 0.01 \, , \;  y_0 \ug [-230 \, , \;54\, , \; 0 \, , \;-342] \, , \; b_0\ug [112 \, , \;41\, , \; 19 \, , \;0]$, iterations $\ug1000$. \label{fs3}}
\end{figure}

\subsubsection{Choice of solutions}\label{solchoice2}
\begin{figure}[H]
 \centering
\hspace{-0.8cm}
\includegraphics[scale=0.5]{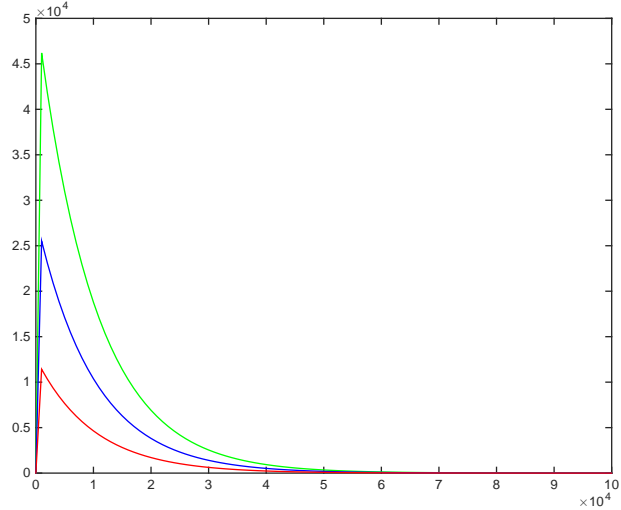}
\caption{We have in the green plot $l_1 \ug -0.0599 \, , \;  l_2 \ug -0.06\, , \;  l_3 \ug -0.01 \, , \;  l_4 \ug -0.0001 $ , in the blue one  $l_1 \ug -0.0199 \, , \;  l_2 \ug -0.1\, , \;  l_3 \ug -0.01 \, , \;  l_4 \ug -0.0001 $ and in the red one $l_1 \ug -0.0399 \, , \;  l_2 \ug -0.04\, , \;  l_3 \ug -0.05 \, , \;  l_4 \ug -0.0001 $. \label{IRexso} }
\end{figure}

Like in the First order case, we are allowed to choose a suitable set of solutions and then find the parameters for our model (Section \ref{findparam}). In this case we have four solutions related to $\t$ by the relation $2\t  \ug -(\l_1 + \l_2 + \l_3 + \l_4)$. Also in this case we need a solution close to $0$ to allow memorization. It is also useful not to choose another one or two little (w.r.t. $\t$) solutions since this makes the $g$ grow too much (Fig.\ref{IRexso}).

% % %% % %  % %  % %\input ./Subsections/tau
\subsection{Sampling-step $\q$, Parameter $\t$ and number of Impulses}\label{taustudy}

In our first experiments we consider a totally supervised Training set, where the examples are equally spaced both in time and space. The first example comes at $t_1 \ug \q$, then the first supervision came at $\frac{3}{2} \q$ (see Section \ref{discretization}) and the system receives an impulse. The next example comes $\q$ seconds after the first and so on. Since the memory is related to the saturation time of impulse response, the learning process of the system embraces all the examples appearing in the interval of time before this saturation. This means that the system has to be build so as the saturation time comes after the whole Training Set has been seen more times. Further more, since the functional in (\ref{S}) contain the term $e^{\t t}$, the parameter $\t$ has to be such that $e^{\t t_0} \ug 1$ is not too much smaller than $e^{\t t_l} $ ($t_l$ instant at which the last example $u_l$ comes). Another important characteristic to take in account  is the delay the Impulse Response. If the supervisions (and then the impulses themselves) are too frequent, there is an accumulation of this delay that could cause instability. For these reasons it is important to study the behavior of the system w.r.t. the rate between $\t$ and $\q$, with some adjustment allowed by the other parameters. In this section its convenient to restrict the analysis to the second order differential operator by managing the solutions of the differential equations (\ref{diff4}). The parameter $\t$ is directly related to the roots $\l$ by the relation 
$$2\t  \ug -(\l_1 + \l_2 + \l_3 + \l_4).$$
This allow us to choose a suitable $\t$ for our model by the solutions, then find the other parameters of the model to optimize the behavior (see Section \ref{findparam}). One solution close to $0$ gives memory to our system. So we choose $\l_1$ small enough to give sufficient memory w.r.t. data, from now on we fix $\l_1 \ug -1 \times 10^{-8}$. We split $\t - \l_1$ roughly equally among the others solutions, since in this way we have a quicker response with a smaller maximum (see Section \ref{findparam}), we assume they are respectively the 60,65 and 75\% of $2\t$. Because of this choice on the parameters, in the following figures we plot at the top both the behavior of function $g$ near the origin and its global trend. We also pose $\eta \ug \g / \mu$. At the bottom we plot the last five iteration on the Training Set, to better see how the oscillation at the steady state depends on the data set. In the label we specify SO to indicate we are dealing with the second order operator.

\begin{figure}[H]
\hspace{-0.8cm}
\includegraphics[scale=0.7]{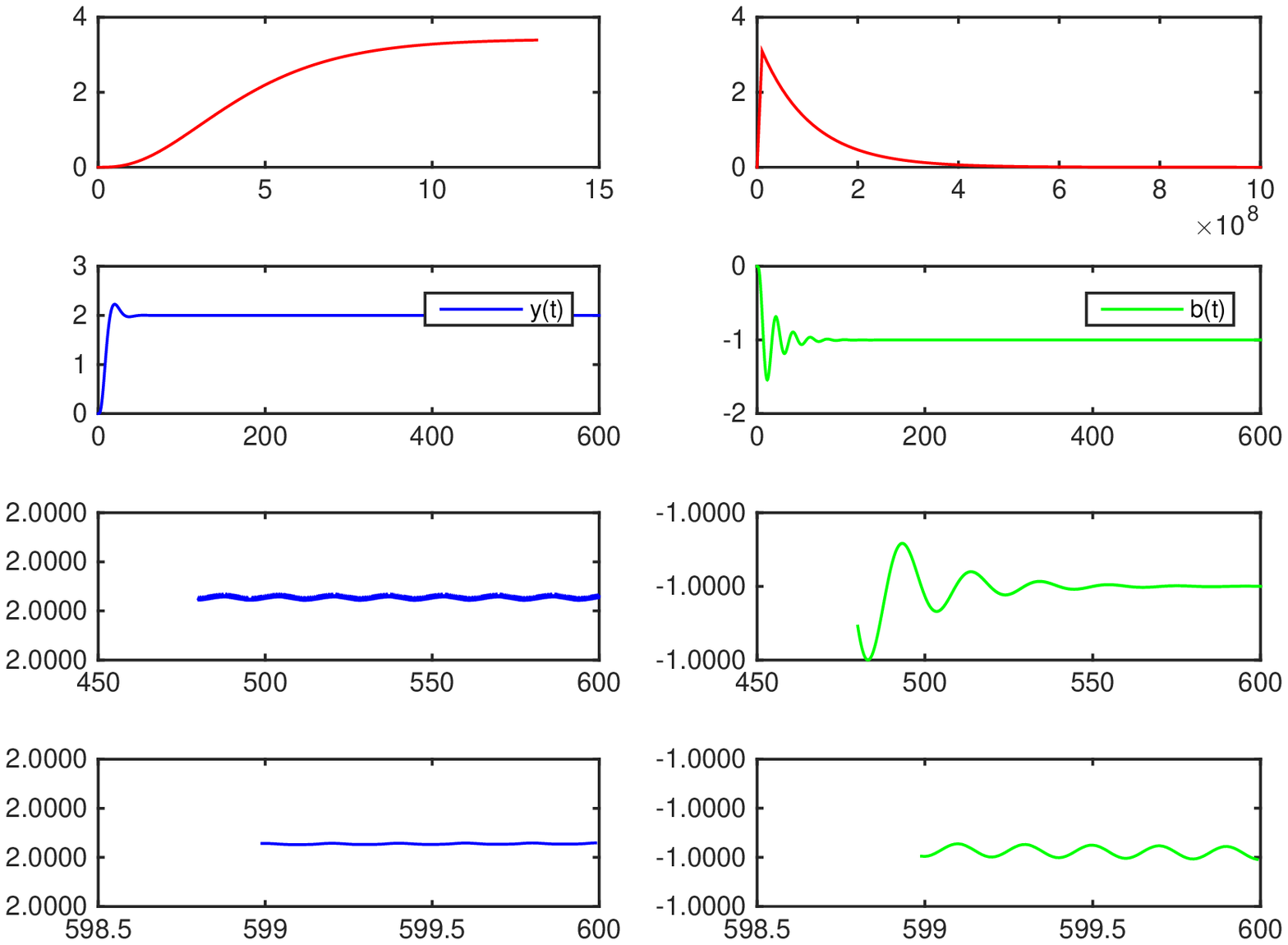}
\caption{SO: $\; \eta \ug 0.001 \, , \; \t \ug 1  \, , \; \q \ug 0.01$, iterations $\ug 3000$. \label{Tau2}}
\end{figure}

As a first experiment we try to better understand the oscillatory behavior when the system reach the steady state. For start we choose our models. We start with the same Training set with $l \ug 20$ and $\q \ug 0.01$, $\t \ug 1$ produce $e^{\t t_l}  \ug 1.22$, using the parameters reported in Fig.\ref{Tau2} and referring to this configuration as the \emph{low} dissipation one.

\begin{figure}[H]
\hspace{-0.8cm}
\includegraphics[scale=0.7]{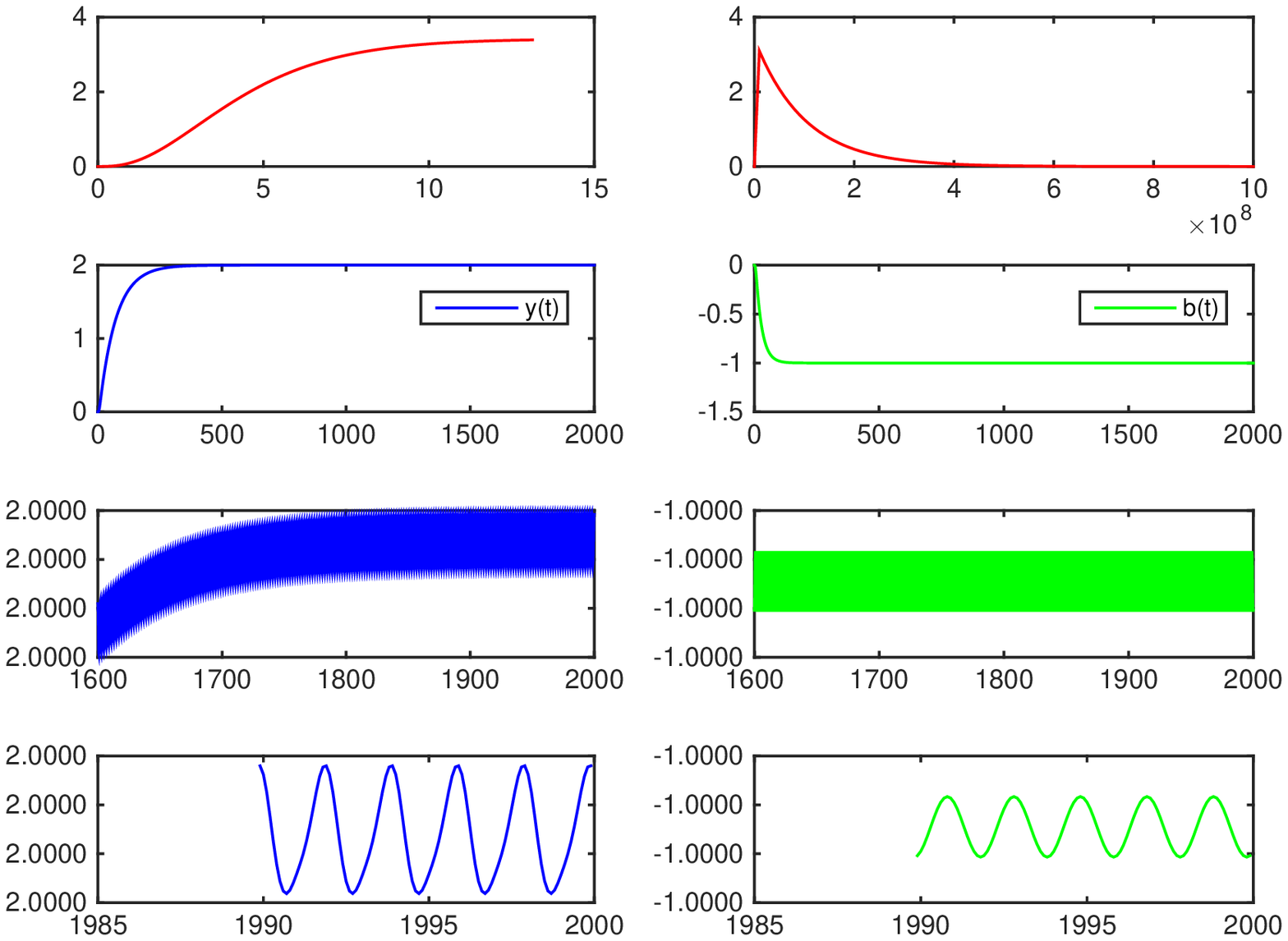}
\caption{SO: $\; \eta \ug 0.001 \, , \; \t \ug 1  \, , \; \q \ug 0.1$, iterations $\ug 2000$. \label{Tau4}}
\end{figure}

When we increase the parameter $\q \ug 0.1$, also the period $T$  and the oscillation period have the same increment, as shown in Fig.\ref{Tau4}.

\begin{figure}[H]
\hspace{-0.8cm}
\includegraphics[scale=0.7]{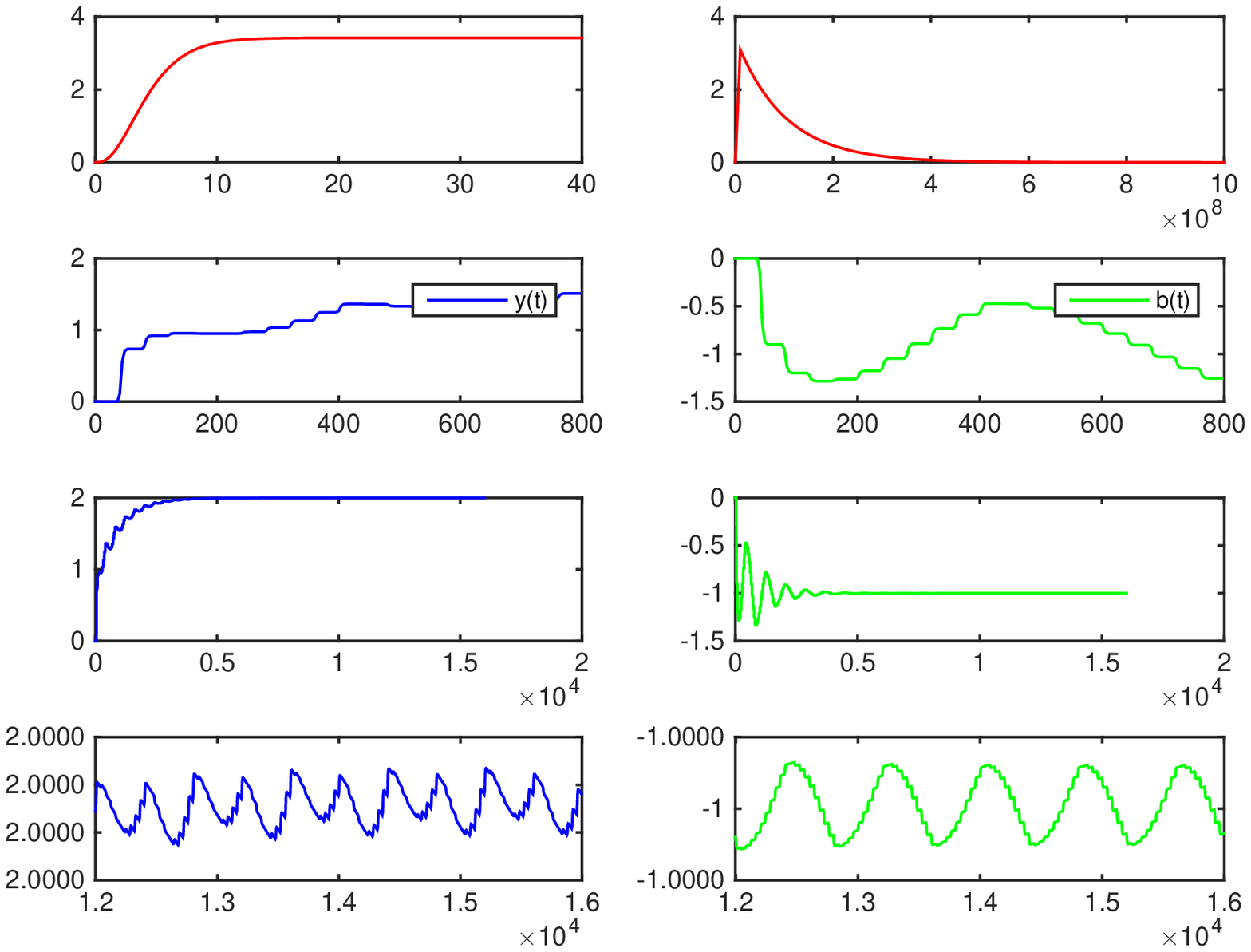}
\caption{SO: $\; \eta \ug 0.1 \, , \; \t \ug 1  \, , \; \q \ug 40$, iterations $\ug 20$. \label{Tau9}}
\end{figure}

Now we show the behavior of the system when we reproduce a situation similar to the classic \emph{Stochastic Gradient Descent}. Since the memory of the system is very long w.r.t. data (because of $\l_1$), the function $g$ is almost constant once it reach its maximum. If the examples are far enough to allow the system to respond between two of them (i.e. $g$ reach its maximum), we reply the gradient descent algorithm. Each examples modify the weights with a term related to the gradient calculated at the previous instant of time. As already said, the period both depends on $\q$ and $\t$, so that we can arbitrarily fix $\t = 1$ and enlarge $\q=40$ so as to obtain the desired configuration with $e^{\t t_l} \ug 10^{173}$. The outcome of this \emph{high} dissipation setting is showed in Fig.\ref{Tau9}. In the remainder of this Section we can see at the top of the figures $g$ again, then the behavior of the two weights $y$ (blue) and $b$ (green) after the first iteration on the Training Set, the total trend and the last 5 iterations at the bottom.

% % % % % %
\section{First application on ANNs}

As first simple practical application we try some experiment in the optimization of a simply ANN. We use a network with one hidden layer and an output layer, the identity as output function and the \emph{rectifier} function:
$$
f(x) = 
\left\{
\begin{array}{ll}
x & \mbox{ if }  x \! > \! 0 \\
0 & \mbox{ otherwise} \\
\end{array}
\right.
$$
 as activation function. In this model we have simply to extend the updating formulas for the weights $y,\;b$ to the weights of the two layers. In this first application we try different setting of the parameters, with different number of units and for both the first and the second order differential operator. In the next list of experiments, we refer to some results obtained by use $\t \ug 1$ and 20 of units in the hidden layer, $\q$ varying so as to guarantee a good value of $e^{\t \q l}$, $\eta$ has been changed to guarantee the best fitting, in the second order differential operator case.

\subsection{One dimension functions}

\begin{figure}[H]
\centering
\includegraphics[scale=0.4]{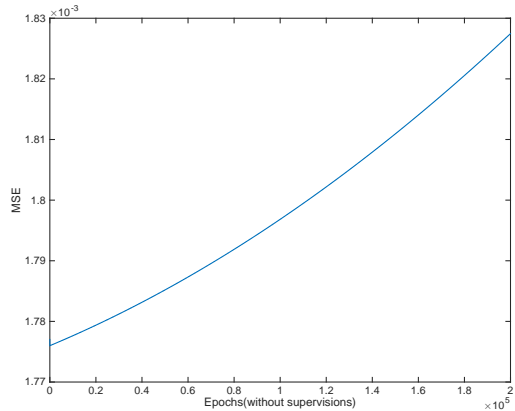}
\caption{Trend of MSE in $2 \! \cdot \! 10^{5}$ iterations without supervisions in regression task. Parameters of the system: $\eta \ug 10^{-4} \, , \; \t \ug 1 \, , \; \q \ug 0.01$, Units: 20. \label{ANN1D1}}
\end{figure}

We first attack the practical model of the first sections, i.e. a regression task on  a Set containing $100$ points $u_k \inn [-1,1]$, which are sorted and equally spaced. All the points are labeled with the target $\y_k \ug 2 \! \cdot \! u_k -1$, but only $10$ points give supervision. We have the MSE$= 1.77 \! \cdot \! 10^{-3}$ after $2 \! \cdot \! 10^{4}$ iterations. In Fig.\ref{ANN1D1} we can see the trend of MSE in other $2 \! \cdot \! 10^{5}$ epochs if we turn off the supervisions (only labeled points for MSE evaluation).

%In Table \ref{tabANN1Dr1} we report some results in term of MSE for different targets and settings. MSE on.

%\input ./Table/TabANN1Dr1

We then try the same parameters for a classification task on the same set. We assign the target class \emph{true} ($\by_k \ug [1\; 0]'$) to the points in $[-0.5,\, 0.5]$ and the class \emph{false} ($\by_k \ug [0\; 1]'$) to the others. Again we use only $10$ points for supervision. After $5 \! \cdot \! 10^{4}$ iterations we have MSE$= 0.03$ and Accuracy$=0.97$. Again we try to turn of the supervisions and go on with agent for other $2 \! \cdot \! 10^{5}$ epochs. Both MSE and accuracy remain almost the same.

% The results are in Table \ref{tabANN1Dc1} 

%\input ./Table/TabANN1Dc1

\subsection{Two dimensions functions}

 We choose our point in $[-1,1] \! \times \! [-1,1] $ . We use two different trajectories to cover the Training set, a spiral and a flower. We obtain the points of the spiral as:

$$
\u(t) = 
\left\{
\begin{array}{l}
(t/100) \cos(t) \\%\frac{t}{20} 
(t/100) \sin(t)\\%\frac{t}{20} \sin(t)  
\end{array}
\right.
$$
whereas the flower trajectory is obtained by
$$
\u(t) = 
\left\{
\begin{array}{l}
  \cos(10 t) \cdot \cos(t) \\
 \cos(10 t) \cdot \sin(t)  \\
\end{array}
\right.
$$

We take $100$ supervised points coming from each trajectories with $t \ug {1 , ...,100}$ ($26$ and $40$ for the class true, respectively for the flower and spiral trajectory). The points in $\left\{ (x,y) \inn \R^2 : |x| + |y| \leq 0.5 \right\} $ represent the class \emph{true}, the others are false. We divide our experiments in two different phases. In the first phase, we train the network with the supervised points for $10^5$ iterations on the set obtained with one trajectory. In the second phase (validation) we check the performance of the system after some epochs without supervisions. We evaluate the performance on the two sets coming from different trajectories and on set obtained by an equally spaced grid of $[-0.5,0.5] \! \times \! [-0.5,0.5] $ containing $100$ examples, equally split in true and false label. The results are in Table \ref{ANN2D1}.

%spiral train 1e5  Train: F:MSE=0.10;A=0.93 S:MSE=0.88;A=0.07 G: MSE=0.06;A=0.80 ;
%			 		V1e5: F:MSE=0.04;A=0.96 S:MSE=0.03;A=0.99 G: MSE=0.05;A=0.91
%			 		V5e5: F:MSE=0.04;A=0.96 S:MSE=0.03;A=0.99 G: MSE=0.05;A=0.91
%flower train 1e5  Train: F:MSE=0.02;A=0.97  S:MSE=0.04;A=0.97 G: MSE=0.06;A=0.97 ;  
%					V1e5: F:MSE=0.22; A=0.74  S:MSE=0.20;A=0.60 G: MSE=0.15;A=0.60 ; 
%					V5e5: F:MSE=0.22; A=0.74  S:MSE=0.20;A=0.60 G: MSE=0.15;A=0.60 ; 

%New
%spiral train 1e5  Train: F:MSE=0.03;A=0.95 S:MSE=0.03;A=0.96 G: MSE=0.07;A=0.81;
%			 		V1e3: F:MSE=0.10;A=0.71 S:MSE=0.1;A=0.63 G: MSE=0.18;A=0.42;
%			 		V1e3: F:MSE=0.12;A=0.63 S:MSE=0.1;A=0.58 G: MSE=0.19;A=0.40;
%flower train 1e5  Train: F:MSE=0.03;A=0.99 S:MSE=0.03;A=0.98 G: MSE=0.07;A=0.85 ;  
%					V1e3:  F:MSE=0.21;A=0.26 S:MSE=0.20;A=0.40 G: MSE=0.27;A=0.40 ; 
%					V1e3:  F:MSE=0.21;A=0.26 S:MSE=0.20;A=0.40 G: MSE=0.27;A=0.40 ; 

% The results are in Table \ref{tabANN2Dc1} 

%\input ./Table/TabANN2Dc1
\begin{table}[h]
\begin{tiny}
\begin{center}
\resizebox{\textwidth}{!}{%
\begin{tabular}{|c|c|c|c|c|c||c|c|c|c|}
\hline
 & & & & & &\multicolumn{4}{|c|}{Accuracy}  \\
   \cline{7-10}
   & &&&&& Training & \multicolumn{2}{|c|}{Validation} & \\
     & &&&&&  & \multicolumn{2}{|c|}{(no supervisions)} & \\
   Training & & & & && $10^5$  epochs & $10^3$ epochs & $2 \! \cdot \! 10^3$ epochs  & \\
 
 trajectory & $\t$ & $\q$ & $\eta$ & $ e^{\t \q l}$ & Units & $10^4$ sec.  & $10^5$ sec. ($\q \ug 1$) & $10^7$ sec.($\q \ug 100$) & Set  \\

\hline
  		&	&		&			&		&	&	0.96		&	0.71	&	0.58	 &  Spiral  \\
\cline{7-10}
Spiral 	& 1 	& 0.001 	& $10^{-4}$ 	& 1.10 	& 20 &	0.95		&	0.63	&	0.53	 &  Flower  \\
\cline{7-10}
  		&	&		&			&		&	&	0.81		&	0.42	&	0.40	 &  Grid  \\
\hline

  		&	&		&			&		&	&	0.98		&	0.40	&	0.40	 &  Spiral  \\
\cline{7-10}
Flower 	&  	&  	 	& 			& 		& 	 &	0.99		&	0.26	&	0.26	 &  Flower  \\
\cline{7-10}
  		&	&		&			&		&	&	0.85		&	0.40	&	0.40	 &  Grid  \\
\hline

\end{tabular}%
}
\end{center}
\end{tiny}
\caption{First 2D classification results, Training phase with $100$ supervised points covered with different trajectories}\label{ANN2D1}
\end{table}

\subsection{Vowels Classifications}

We record few tracks with the sequential pronunciation of the five Italian vowels. We process the files with Matlab and take the auditory spectra coefficients coming from RASTA PLP algorithm. We obtain a sampling of the tracks with points in a $40$ dimensions features space. Set 1 is obtained from a $20$ seconds track with sequential pronunciation of the vowels($2053$ samples, $700$ labeled). Set 2 ($2144$ samples) is obtained from a $20$ seconds track with sequential pronunciation of the vowels repeated in time  ($600$ labeled points). Set 3 is obtained from 5 tracks, each containing the pronunciation of a vowel ($14934$ labeled points). We carry on the experiments with the same approach of the previous section. In the first phase we train the net with a few supervised samples. Again, in the second phase we turn off the supervisions and study the performance of the system as time goes by. After some epochs, we evaluate the agent on each set.  The results are in Table \ref{ANNV1}.

%U20
%Supervisions: 30 		Train1e5: Set1 MSE:0.0.004 A:1   Set2 MSE:0.03 A:0.82  Set3 MSE:0.04 A:0.84  	
%						V1e3 Set1 MSE:0.005 A:1   Set2 MSE:0.03 A:0.82  Set3 MSE: 0.04 A:0.84 	
%						V1e3 Set1 MSE:0.02 A:0.95   Set2 MSE:0.05 A:0.64  Set3 MSE:0.04 A:0.77  
%Supervisions: 60 		Train1e5: Set1 MSE:0.004 A:1   Set2 MSE:0.02 A:0.90  Set3 MSE:0.03 A:0.87  	
%						V1e3 Set1 MSE: 0.004 A:1   Set2 MSE: 0.02A:0.91  Set3 MSE: 0.03 A:87 	
%						V1e3 Set1 MSE:0.01 A:1   Set2 MSE:0.028 A:0.93  Set3 MSE:0.03 A:0.87
%Supervisions: 90 		Train1e5: Set1 MSE:0.004 A:1   Set2 MSE:0.014 A:0.95  Set3 MSE:0.019 A: 0.96 	
%						V1e3 Set1 MSE:0.004 A:1   Set2 MSE:0.014 A:0.95  Set3 MSE:0.019 A: 0.96	
%						V1e3 Set1 MSE: 0.015 A:1   Set2 MSE: 0.02 A:0.93  Set3 MSE:0.02 A:0.95
%U100
%Supervisions: 30 		Train1e5: Set1 MSE:0.004 A:1   Set2 MSE:0.03 A:0.79  Set3 MSE:0.04 A:0.83 
%					 	V1e3 Set1 MSE:0.005 A:1   Set2 MSE:0.035 A:0.78  Set3 MSE: 0.04 A:0.83
%						V1e3 Set1 MSE: A:   Set2 MSE: A:  Set3 MSE: A:
%Supervisions: 60 		Train1e5: Set1 MSE:0.004 A:1   Set2 MSE:0.018 A:0.94  Set3 MSE:0.029 A:0.90  	
%						V1e3 Set1 MSE:0.004 A:1   Set2 MSE:0.018 A:0.94  Set3 MSE: 0.02 A:0.90 	
%						V1e3 Set1 MSE:0.01 A:1   Set2 MSE:0.02 A:0.94  Set3 MSE: 0.03 A:0.90 
%Supervisions: 90 		Train1e5: Set1 MSE: 0.003 A:1   Set2 MSE:0.014 A: 0.95 Set3 MSE:0.017 A:0.96  	
%						V1e3 Set1 MSE:0.003 A:1   Set2 MSE:0.014 A:0.95  Set3 MSE:0.011 A: 0.96	
%						V1e3 Set1 MSE:0.012 A:1   Set2 MSE:0.02 A:0.95  Set3 MSE:0.016 A:0.96

% MC% Train 
\begin{table}[h]
\begin{scriptsize}
\begin{center}
\resizebox{\textwidth}{!}{%
\begin{tabular}{|c|c|c|c|c|c||c|c|c|c|}
\hline
 & & & & & &\multicolumn{4}{|c|}{Accuracy}  \\
   \cline{7-10}
   & &&&&& Training & \multicolumn{2}{|c|}{Validation} & \\
     & &&&&&  & \multicolumn{2}{|c|}{(no supervisions)} & \\
   & & & & && $10^5$  epochs & $10^3$ epochs & $2 \! \cdot \! 10^3$ epochs  & \\
    & & & & && $10^5$  epochs & $ 2.053 \! \cdot \! 10^6$ sec. & $2.053 \! \cdot \!10^7$ sec. & \\
Supervisions & $\t$ & $\q$ & $\eta$ & $ e^{\t \q l}$ & Units &  &   ($\q \ug 1$) & ($\q \ug 10$) & Set  \\

\hline
\hline
  			&	&		&			&		&	&	1		&	1	&	0.99	 &  Set1  \\
\cline{7-10}
30 from Set1 	& 1 	& 0.01 	& $10^{-5}$ 	& 1.34	& 20 &	0.82		&	0.82	&	0.72	 &  Set2  \\
\cline{7-10}
  			&	&		&			&		&	&	0.84		&	0.84	&	0.82	 &  Set3  \\
\hline
\cline{7-10}
  			&	&		&			&		&	&	1		&	1	&	0.95	 &  Set1  \\
\cline{7-10}
		 	&  	& 	 	& 		 	& 		& 100&	0.79		&	0.78	&	0.64	 &  Set2  \\
\cline{7-10}
  			&	&		&			&		&	&	0.83		&	0.83	&	0.77	 &  Set3  \\
\hline
\hline
 30 from Set1	&	&		&			&		&	&	1		&	1	&	1	 &  Set1  \\
\cline{7-10}
30 from Set2 	& 1 	& 0.005 	& $10^{-5}$ 	& 1.34	& 20 &	0.90		&	0.91	&	0.93	 &  Set2  \\
\cline{7-10}
 			&	&		&			&		&	&	0.87		&	0.87	&	0.87	 &  Set3  \\
\hline
\cline{7-10}
  			&	&		&			&		&	&	1		&	1	&	1	 &  Set1  \\
\cline{7-10}
		 	&  	& 	 	& 		 	& 		& 100&	0.94		&	0.94	&	0.94	 &  Set2  \\
\cline{7-10}
  			&	&		&			&		&	&	0.90		&	0.90	&	0.90	 &  Set3  \\
\hline
\hline
  30 from Set1	&	&		&			&		&	&	1		&	1	&	1	 &  Set1  \\
\cline{7-10}
30 from Set2 	& 1 	& 0.003 	& $10^{-5}$ 	& 1.31	& 20 &	0.95		&	0.95	&	0.93	 &  Set2  \\
\cline{7-10}
30 from Set3	&	&		&			&		&	&	0.96		&	0.96	&	0.95	 &  Set3  \\
\hline
\cline{7-10}
  			&	&		&			&		&	&	1		&	1	&	1	 &  Set1  \\
\cline{7-10}
		 	&  	& 	 	& 		 	& 		& 100&	0.95		&	0.95	&	0.95	 &  Set2  \\
\cline{7-10}
  			&	&		&			&		&	&	0.96		&	0.96	&	0.96	 &  Set3  \\
\hline			
\end{tabular}%
}
\end{center}
\end{scriptsize}
\caption{First  Vowels classification results}\label{ANNV1}
\end{table}

% % % % % % CONCLUSION
\section{Conclusions}

As already said, the studied applications are not the perfect suit of our theory. However, the positive results showed could strengthen our hypothesis and help to better understand the meaning of the different aspects. This made us look for a deeper analysis from many theoretical points of view. We are talking about study others differential operators, cost functionals and forms of the function $f$. Simultaneously, we would like to investigate the behavior of the current model in applications in which a manifold regularization in time plays an fundamental role, as in Computer Vision problems.

\section{Appendix}\label{appendix}%{\Large{{\bf \noindent Appendice A.}} {\Large{{\bf Scripts con Maple ed esperimenti}}}}
%\addcontentsline{toc}{chapter}{Appendice A. Esperimenti e scripts con Maple}

\subsection{General solution and coefficients}

In this section we report some practical calculations and assumptions to solve the differential equation of our theoretical framework. The notation is finalized to a practical general implementation. 

We can write the general form of characteristic polynomial as:

\begin{equation}\label{cp}
\b_n \l^n + ... + \b_1 \l + \b_0  =0 
\end{equation}

Then we have a set of $J$ solution $\l_j$ each with they multiplicity $r_j$. The Laplace Transform lead to:

\begin{equation}\label{Gs}
G(s)=\frac{1}{\sum_{q=0}^n \b_q \l^q}   = \sum_{j=1}^J \sum_{i=1}^{r_j} \frac{c_{ji}}{{(s- \l_j)}^{r_i}}
\end{equation}

Where $c_{ji}$ are constants such that:

\begin{equation}\label{Ceq}
 \sum_{j=1}^J \sum_{i=1}^{r_j} c_{ji}{(s- \l_j)}^{r_j-i}( \prod_{\begin{array}{c}\scriptstyle k=1\\ \scriptstyle k \neq i \end{array}}^J {(s- \l_k)}^{r_k}) = 1
\end{equation}

For practical issue we pose

\begin{equation} 
\begin{array}{cccccccccccccc}
\Lambda^{j,j} & = & [  & \l_1 & \cdots & \l1 & \cdots & \l_J &  \cdots & \l_J & \l_j & \cdots & \l_j &   ] \\
R^{j,j} & = & [   & 1 & \cdots & r1 & \cdots & 1 &  \cdots & r_J &1& \cdots & r_j &  ] \\
\Lambda^{j,q} & = & [ &   \l_1 & \cdots & \l1 & \cdots&  \l_J &  \cdots & \l_J & \l_j&  \cdots & \l_q &   ] \\
R^{j,q} & = & [  &  1 & \cdots & r1 & \cdots & 1 &  \cdots&  r_J& 1& \cdots & r_q  & ] \; , \, q \leq j\\
I_n & = & [ & 1 &\cdots &n &]&&&&&&&
\end{array}
\end{equation}

 To simplify the notation we pose $R \ug R^{J,J}, \, \Lambda \ug \Lambda^{J,J}$. If we carry out the summation in (\ref{Gs})(with the new notation), pose $n_j \ug n \! - \!R_j$ we find that each $c_{ji}$ multiply a factor:
\begin{equation}
c_{ji} \left[   \sum_{k=0}^{n_j} s^{n_j-k} \left( \sum_{i \! \in \! C_{n_j,k}(I_{n_j}) }  (  \prod_{p=i_1}^{i_{n_j}} -\Lambda^{j,i}_p  )   \right)     \right] = c_{ji}   \sum_{k=0}^{n_j} s^{n_j-k} A_{l,j+i} \, , \; l=n_j-k+1
\end{equation}
then we can determine $C=[ c_{11} \cdots c_{1r_1} \cdots c_{J1} \cdots  c_{J r_J}]'$ from the system $A C=b$ , $b=[ 1 0 \, \cdots \, 0 ]'$ and $A \inn \R^{n,n}$ with:

\begin{equation}
A_{l,j+i}=\left\{
\begin{array}{ll}
\sum_{i \! \in \! C_{n_j,k}(I_{n_j}) }  (  \prod_{p=i_1}^{i_{n_j}} -\Lambda^{j,i}_p  )  & \mbox{ if } 1\leq l \leq n_j+1\, , \; k \ug n_j +1 -l \\
0 & n_j+1 < l \leq n 
\end{array}
\right.
\end{equation}

The general solution is of the form:

\begin{equation}\label{gs}
g(t)=\sum_{j=1}^J C_{j} t^{R_j-1} e^{\Lambda_j t}
\end{equation}

When we have a complex solution $\l_j=\a+i\b$, also its conjugate $\l_p=\bar{\l_j}=\a -i\b$ is present and also the relatives constants are such that $C_p=\bar{C_j}=\a_c -i \b_c$. We have in the solution :
$$
\cdots + C_j \cdot e^{\a t}\left( \cos \b t + i \sin \b t \right) + C_p \cdot e^{\a t}\left( \cos (-\b t) + i \sin (-\b t) \right) + \cdots 
$$
$$
\cdots + C_j \cdot e^{\a t}\left( \cos \b t + i \sin \b t \right) + C_p \cdot e^{\a t}\left( \cos (\b t) - i \sin (\b t) \right) + \cdots 
$$
\begin{equation}\label{comp}
\cdots + (C_j+C_p)\cdot e^{\a t}\left( \cos \b t \right) +(C_j-C_p)\cdot e^{\a t}\left( i \sin \b t \right) + \cdots   
\end{equation}
and since
\begin{eqnarray*}
(C_j+C_p) & = & \a_c + i\b_c +\a_c -i\b_c \\
(C_j-C_p) & = & \a_c + i\b_c -\a_c +i\b_c \\
\end{eqnarray*}
the (\ref{comp}) becomes
$$
\cdots + 2 \a_c \cdot e^{\a t}\left( \cos \b t \right) + 2 i \b_c\cdot e^{\a t}\left( i \sin \b t \right) + \cdots   
$$
\begin{equation}\label{comp2}
\cdots + 2 \a_c \cdot e^{\a t}\left( \cos \b t \right) - 2  \b_c\cdot e^{\a t}\left(  \sin \b t \right) + \cdots  
\end{equation}
since this is the contribution of two solution with the same $real$ and $imaginary$ parts, we can write (\ref{gs}) as

\begin{equation}\label{gcs}
g(t)=\sum_{j=1}^J  t^{R_j-1} e^{\Re(\Lambda_j) t} \left(\Re(C_{j})(\cos(\Im(\Lambda_j)t))-\Im(C_{j}) (\sin(\Im(\Lambda_j)t)) \right)
\end{equation}

also the solution to the homogeneous equation is of the form

\begin{equation}\label{gso}
y^o(t)=\sum_{j=1}^J K_{j}t^{R_j-1}e^{\Lambda_j t}
\end{equation}

where $K_{j}$ are determined by imposing the Initial Conditions given for $y^o(t)$, that is $Y_0=[y^o(0) \cdots {y^o}^{(n-1)}(0)]'$. Since
\begin{equation}\label{dos}
{y^o}^{(d)}(0)=\sum_{j=1}^J (d+2-R_j)^+K_{j}\Lambda_j^{(d+1-R_j)^+}
\end{equation}
we can find $K=[K_1 \cdots K_n]'$ by solving the system $MK=Y_0$ where:
\begin{equation}\label{dos}
M_{vj}=(v+1-R_j)^+\Lambda_j^{(v-R_j)^+}
\end{equation}

and exactly as in the case of $g(t)$ we can write 
\begin{equation}\label{gcs}
y(t)=\sum_{j=1}^J  t^{R_j-1} e^{\Re(\Lambda_j) t} \left(\Re(K_{j})(\cos(\Im(\Lambda_j)t))-\Im(K_{j}) (\sin(\Im(\Lambda_j)t)) \right)
\end{equation}

\subsection{From solution to parameters}\label{findparam}

In section \ref{2ndOrd} we saw that for the second order case convergence depends not only on $\g$, but also on the kind of the solutions of the characteristic polynomial, i.e. on $\b_j$. Moreover, the most important parameter of our model is $\t$, which allow to choose the memory width of the system. This memory is related to the function $\psi(t) \ug e^{\t t}$, which represent the weight that the model assigns to each samples as the time goes by. The smaller is $\t$( but always $>0$) the bigger is the memory of our model. We are interested in find suitable values of $\a_j$ which allow convergence when $\t$ is small. In practice, we can build our model only by the solutions of the characteristic poly, since the parameters influence the updating formulas (\ref{upfo}),(\ref{fbp2}) only by the last $\a_j$, which can be absorbed in the term $\mu$. Then we can choose directly suitable solutions, and verify that they are related to meaningful values of parameters. So, starting from the solutions (chosen in a way to have the desired $\t$), is easy to find the coefficients of the characteristic poly and then find the rate between the $\a_j$ that generate the model.

\subsubsection{First Order}

If we have the solutions $\l_1 , \, \l_2$ the characteristic poly of (\ref{diff}) is 
$$
\l^2 + \t \l + \b = \left( \l - \l_1 \right) \left( \l - \l_2 \right) .
$$

So the value of $\t$ is given by $\theta = - \left( \l_2 + \l_1 \right) $ and the rate $\frac{\a_0 }{ \a_1}$ is computable from $\b$. That is, if we pose ${\nu}=\frac{\a_0}{\a_1}$, we can find suitable value of $\a_j$ from the solutions of

 \begin{equation}\label{nu}
 \nu^2 - \t \nu + \b = 0.
 \end{equation}

\subsubsection{Second Order}

In this case the characteristic poly is $\l^4 + \b_3 \l^3+ \b_2 \l^2 + \b_1 \l + \b_0  =0$. The coefficient $\b_3$ is still given by the opposite of the summation among the solutions, and since $\b_3 \ug 2\t$ it is still easy to set $\t$. To find the rates among $\a_j$ is convenient to work with ${\nu}_0=\frac{\a_0}{\a_2}$ and ${\nu}_1=\frac{\a_1}{\a_2}$ so that:

 \begin{equation}\label{b0}
 \b_0 \!  \ug  \frac{\a_0 \a_2 \t^2 - \a_0 \a_1 \t  + \a_0 ^2}{a_2^2}   \ug  {\nu}_0 \t^2 - {\nu}_0 {\nu}_1 \t + {\nu}_0^2 
 \end{equation}
 
 \begin{equation}\label{b1}
  \b_1 \!  \ug  \frac{\a_1 \a_2 \t^2 + (2 \a_0 \a_2 - \a_1^2)\t}{\a_2^2}  \ug  {\nu}_1 \t^2 + 2 {\nu}_0  \t - {\nu}_1^2 \t 
  \end{equation}
  
  \begin{equation}\label{b2}
 \b_2 \!  \ug  \frac{\a_2^2 \t^2 + \a_1 \a_2 \t + 2 \a_0 \a_2 - \a_1^2}{\a_2^2} \ug  \t^2 + {\nu}_1 \t + 2 {\nu}_0 - {\nu}_1^2 
 \end{equation}

% \begin{equation}\label{coeff4redux}
 %\begin{array}{rclcl}
 %\b_0 \! & \ug & \frac{\a_0 \a_2 \t^2 - \a_0 \a_1 \t  + \a_0 ^2}{a_2^2}  & \ug & {\nu}_0 \t^2 - {\nu}_0 {\nu}_1 \t + {\nu}_0^2 \\
  %\b_1 \! & \ug & \frac{\a_1 \a_2 \t^2 + (2 \a_0 \a_2 - \a_1^2)\t}{\a_2^2} & \ug & {\nu}_1 \t^2 + 2 {\nu}_0  \t - {\nu}_1^2 \t \\
 %\b_2 \! & \ug & \frac{\a_2^2 \t^2 + \a_1 \a_2 \t + 2 \a_0 \a_2 - \a_1^2}{\a_2^2}& \ug & \t^2 + {\nu}_1 \t + 2 {\nu}_0 - {\nu}_1^2 \\
% \b_3 \! & \ug & 2\t & &
 %\end{array}
 %\end{equation}

From the equation (\ref{b2}) we can find the relation among the coefficients:
 \begin{equation}
 \b_1 \ug \frac{\b_3 \b_2}{2}-\frac{\b_3^3}{8}
  \end{equation}
  
from equation (\ref{b1}) we find
 \begin{equation}
\nu_0 \ug \frac{1}{2\t}\left( \b_1+ \nu_1^2 \t - \nu_1 \t^2 \right)
  \end{equation}
  
  and from (\ref{b0}) we have 
 \begin{equation}\label{nu1}
\nu_1^4 -4 \t \nu_1^3 + \nu_1^2 \left( 5 \t^2 + 2 \b_1 / \t \right) + \nu_1 \left( -2 \t^3 -4 \b_1  \right) + \left( 2 \t \b_1+ \b_1^2 / \t^2 -4 \b_0 \right) \ug 0.
  \end{equation}
  
  Once we find the solutions of (\ref{nu1}) we can choose a suitable one and find $\nu_0$, and then have an idea of which $\a_j$ generate our model. For practical reasons, since $\b_0 \ug \l_1 \, \l_2 \, \l_3 \, \l_4$, by using (\ref{b0}) we can find $\a_1$ by posing $\a_0 \ug \a_2 \ug 1 $ in (\ref{b0}):
  
\begin{equation}
\a_1 \ug \nu_1 \ug \frac{\nu_0 \, \t^2 + \nu_0^2 - \b_0}{\nu_0 \, \t}.
\end{equation}

In our study on the model parameters, we see that is convenient to have one solution close to zero (which guarantees memory to the system). We can choose $\l_1 \ug 1/ a$ (where $a$ is a parameters which represent the memory of the system, since the saturation time is proportional to $a$). Since the modulus of the other solutions determine the shape of the Impulse response, is convenient to write the solutions as:
 \begin{equation}\label{solut}
\begin{array}{ccl}
\l_1 & \ug & c_1 \, \t \ug 1 /a \\
\l_2 & \ug & c_2 \, \t \\
\l_3 & \ug & c_3 \, \t \\
\l_4 & \ug & c_4 \, \t \\
\end{array}
\end{equation}

where $\sum c_j \ug 2$ and $c_2\, , \; c_3\, , \; c_4$ as to be similar among them (but not too much to avoid numerical error) to give a quick Impulsive Response. 

Now we are allowed to choose suitable solutions w.r.t. the model and then find the value of $\g / \mu$ which guarantee convergence and the best fitting performance, both for first and second order.

\subsection{From continuos to discrete model}\label{discretization}

In the practical implementation is not convenient to use the continuos updating formulas (\ref{upfo}),(\ref{fbp2}), since we have to store too much value of the gradient, that is, the wider is memory of the system, the bigger is the number of elements that we have to remember. For this reason, is convenient to use a discretization of the system. Indeed, when we have a linear differential equation of order bigger than one, we can transform it in a system of the same order with only linear differential equations of order one. In our case in
$$
D^4y + \b_3 D^3 y + \b_2 D^2 y + \b_1 D y + \b_0  y + \eta \sum_{k =1}^l \z_k \cdot \delta (t-t_k) \ug 0
$$
we can substitute $y_0 \ug y , \, y_1 \ug D y , ...$ and posing $u(t) \ug \eta \sum_{k =1}^l \z_k \cdot \delta (t-t_k) $ so that to have:
\begin{equation}
\medskip
\left\{
\begin{array}{rcl}
y_1 & \ug & y_0' \\
y_2 & \ug & y_1' \\
y_3 & \ug & y_2' \\
 y_{4 }& \ug & - \b_3 y_3 - \b_2  y_2 - \b_1  y_1 - \b_0  y_0 - u(t)
\end{array}
\right.
\medskip
\end{equation}
and then we have the system
\begin{equation}
\medskip
\dot{\y}=
\left[
\begin{array}{c}
y_1 \\
y_2 \\
y_3 \\
y_4 
\end{array}
\right]
=
\left[
\begin{array}{cccc}
   0      &     1     &     0    &      0   \\
  0      &     0     &     1    &      0   \\
  0      &     0     &     0    &      1   \\
- \b_0 & - \b_1 & - \b_2 & - \b_3 \\
\end{array}
\right]
\left[
\begin{array}{c}
y_0 \\
y_1 \\
y_2 \\
y_3 
\end{array}
\right] + \B u = \A \y + \B u
\medskip
\end{equation}
where
\medskip
$$
\A =
\left[
\begin{array}{cccc}
   0      &     1     &     0    &      0   \\
  0      &     0     &     1    &      0   \\
  0      &     0     &     0    &      1   \\
- \b_0 & - \b_1 & - \b_2 & - \b_3 \\
\end{array}
\right] 
\quad
\B = \left[
\begin{array}{r} 0 \\ 0 \\ 0 \\ -1 \\ 
\end{array}
\right]
\medskip
$$
from the Lagrange formula we have
\begin{equation}
\medskip
\y (t) \ug e^{\A(t-t_0)} \y (t_0) + \int_{t_0}^t e^{\A(t-s)} \cdot \B u(s) ds .
\medskip
\end{equation}

When we consider an equally spaced discretization of the time of width $\tau$, a general instant of time is $t = \q K$ and if we assume $t_0 \ug 0$, the general evolution of the system can be computed by
\begin{equation}
\medskip
\y [ K] = \y(\q K)= e^{\A \tau K} \y [0] + \int_{0}^{\q K} e^{\A(\q K-s)} \cdot \B u(s) ds
\medskip
\end{equation}
and at the next step we have 
\begin{eqnarray*}
\medskip
\y [K+1]  & = & e^{\A \tau (K+1)} \y [0] + \int_{0}^{\q (K+1)} e^{\A(\q (K+1)-s)} \cdot \B u(s) ds  \\
              & = & e^{\A \q} \left( e^{\A \tau K} \y [0]  + \int_{0}^{\q K} e^{\A(\q K-s)} \cdot \B u(s) ds \right) +\\
              & + & \int_{\q K}^{\q (K+1)} e^{\A [\q (K+1)-s]} \cdot \B u(s) ds \\
              & = & e^{\A \q} \y [ K]  + \int_{\q K}^{\q (K+1)} e^{\A [ \q (K+1)-s]} \cdot \B u(s) ds \\
              \medskip
\end{eqnarray*}
since $u(t)$ is composed by a summation of impulses we have a summation of integrals in the second term. If we assume that each impulse (i.e. each supervision) is provided in the middle of two step, we have that only the integrals referred to the last  impulse (the one provided at $h \ug \q K + \q /2$) is different from $0$ :
\begin{eqnarray*}
\medskip
\int_{\q K}^{\q (K+1)} e^{\A(\q (K+1)-s)} \cdot \B u(s) ds  & = & \eta \sum_{k =1}^l  \int_{\q K}^{\q (K+1)} e^{\A [ \q (K+1)-s]} \cdot \B \, \z_k \cdot \delta (s-t_k)  \\
										   & = & \eta \int_{\q K}^{\q (K+1)} e^{\A [ \q (K+1)-s]} \cdot \B \, \z_h \cdot \delta [s-(\q K + \q /2)] \\
										   & = & \eta \, e^{\A [\q (K+1)-(\q K + \q /2)]} \cdot \B \, \z_h \\
										   & = & \eta \, e^{\A \q /2 } \cdot \B \, \z_h \\
										   \medskip
\end{eqnarray*}
that is 
\begin{equation}
\medskip
\y [ K+1] = e^{\A \q} \y [ K] +  e^{\A \q /2 } \cdot \B \,  \eta \, \z_h .
\medskip
\end{equation}

\newpage

\bibliographystyle{unsrt}

\bibliography{mybibfile}

\end{document}